\documentclass[fleqn,10pt]{wlscirep}
\usepackage[utf8]{inputenc}
\usepackage[T1]{fontenc}
\usepackage{xurl}
\definecolor{red}{rgb}{200,0,0}
\usepackage{multirow}
\usepackage{adjustbox}
\DeclareCaptionLabelFormat{andtable}{#1~#2  \&  \tablename~\thetable}
\usepackage{floatrow}
\newfloatcommand{capbtabbox}{table}[][\FBwidth]
\usepackage{lineno}
\usepackage{verbatim}
\usepackage{wrapfig}
\usepackage{lineno}

\title{Fractional dynamics foster deep learning of COPD stage prediction}

\author[1]{Chenzhong Yin}
\author[2,3*]{Mihai Udrescu}
\author[1]{Gaurav Gupta}
\author[1]{Mingxi Cheng}
\author[3]{Andrei Lihu}
\author[4]{Lucretia Udrescu}
\author[1*]{Paul Bogdan}
\author[5]{David M Mannino}
\author[6]{Stefan Mihaicuta}
\affil[1]{Ming Hsieh Department of Electrical and Computer Engineering, University of Southern California, Los Angeles, CA, USA}
\affil[2]{Department of Computer and Information Technology, Politehnica University, Timi\c{s}oara, Romania}
\affil[3]{Timi\c{s}oara Institute of Complex Systems, Timi\c{s}oara, Romania}
\affil[4]{Department I -- Drug Analysis, “Victor Babe\c{s}” University of Medicine and Pharmacy Timi\c{s}oara, 2 Eftimie Murgu Sq., 300041 Timişoara, Romania}
\affil[5]{University of Kentucky, College of Medicine, Lexington, KY, USA}
\affil[6]{Department of Pulmonology, “Victor Babe\c{s}” University of Medicine and Pharmacy Timi\c{s}oara, 2 Eftimie Murgu Sq., 300041 Timişoara, Romania}

\affil[*]{Correspondence and requests for materials should be addressed to M.U. (email: mudrescu@cs.upt.ro) and P.B. (email: pbogdan@usc.edu)}

\begin{abstract}
Chronic obstructive pulmonary disease (COPD) is one of the leading causes of death worldwide, usually associated with smoking and environmental occupational exposures. Prior studies have shown that current COPD diagnosis (i.e., spirometry test) can be unreliable because the test can be difficult to do and depends on an adequate effort from the testee and supervision of the testor. Moreover, the extensive early detection and diagnosis of COPD is challenging.
We address the COPD detection problem by constructing two novel COPD physiological signals datasets (4432 medical records from 54 patients in the WestRo COPD dataset and 13824 medical records from 534 patients in the WestRo Porti COPD dataset), demonstrating their complex coupled fractal dynamical characteristics, and performing a rigorous fractional-order dynamics deep learning analysis to diagnose COPD with high accuracy. 
We find that the fractional-order dynamical modeling can extract distinguishing signatures from the physiological signals across patients with all COPD stages---from stage 0 (healthy) to stage 4 (very severe). We exploit these fractional signatures to develop and train a deep neural network that predicts the suspected patients' COPD stages based on the input features (such as thorax breathing effort, respiratory rate, or oxygen saturation levels). We show that our COPD diagnostics method (fractional dynamic deep learning model) achieves a high prediction accuracy (98.66\% $\pm$ 0.45\%) on WestRo COPD dataset and can serve as an excellent and robust alternative to traditional spirometry-based medical diagnosis. Our fractional dynamic deep learning model (FDDLM) for COPD diagnosis also presents high prediction accuracy when validated by a dataset with different physiological signals recorded (i.e., 94.01\% $\pm$ 0.61\% for predicting the COPD stages in the WestRo COPD dataset with the model trained on the WestRo Porti COPD dataset, and 90.13\% $\pm$ 0.89\% for predicting in the WestRo Porti COPD with the model trained on WestRo COPD).
\end{abstract}
\begin{document}

\flushbottom
\maketitle
%
%
\thispagestyle{empty}
\section*{Introduction}

Chronic Obstructive Pulmonary Disease (COPD) is an increasingly prevalent respiratory disorder, which represents a severe impediment for the quality of life \cite{marczak2016and,yoon2011confronting}; it is the third or fourth major cause of death worldwide \cite{lozano2012bin}. Medical practice presents COPD as an inflammatory lung condition consisting of a slow, progressive obstruction of airways that reduces pulmonary capacity \cite{vestbo2013global}. Medical science has not entirely clarified what triggers COPD; nonetheless, scientists indicate the complex interactions between the environmental factors---such as pollution exposure or smoking---and the genetics \cite{agusti2018copd} as likely causes. COPD is not reversible, but early diagnosis creates incentives for achieving a better disease evolution, and an improved patient condition through personalized treatments \cite{agusti2016treatable}.

The Global Initiative for Obstructive Pulmonary Disease (GOLD) defines COPD---based on pulmonary function testing or spirometry---as the ratio between the forced expiratory volume in one second and the forced vital capacity (FEV1/FVC) of $< 0.7$ in a patient with symptoms of dyspnea, chronic cough, and sputum production, with an exposure history to cigarette smoke or biofuels, or occupational particulate matter. The spirometer is a device that measures the lung's volume and air debits, rendered as forced expiratory volume in one second (FEV1), forced vital capacity (FVC), and the ratio between FEV1 and FVC; physicians use these parameters to classify patients in one of the following COPD stages: 1--Mild, 2--Moderate, 3--Severe, and 4--Very Severe. The almost unanimously accepted classification methodology is the COPD Gold Standard \cite{celli2004standards,kerstjens2004gold}, although there are some differences in applying it \cite{miravitlles2016review}. Unfortunately, early COPD detection and diagnosis are challenging at the population level because relevant clinical signs are hard to detect in the early phases. When suspected, patients are ordinarily subjected to pulmonary function tests (i.e., spirometry) and mostly diagnosed when they are already in stages 2--4; {thus, designing therapies to improve the disease trajectory becomes difficult \cite{goossens2014does}.} Another problem with spirometry is that it does not always render reliable results, mainly when not performed in a specialized pulmonary center \cite{velickovski2018automated}. Nonetheless, the fact that COPD has become a global threat \cite{marczak2016and,yoon2011confronting} further emphasizes the importance of decentralizing diagnosis, meaning that finding innovative methods to diagnose COPD outside respiratory medicine centers becomes paramount.  Recent medical research suggests that personalized medicine could improve COPD diagnosis \cite{agusti2016treatable}. One approach to COPD personalized care is identifying patient phenotypes based on comorbidities, simple clinical, and anthropometric data (e.g., age, body-mass index, smoker status). To this end, the medical practice uses two questionnaires to evaluate symptoms and evaluate the severity of the disease, namely COPD Assessment Test (CAT) and Medical Research Council Breathlessness Scale (MRC) \cite{dodd2011copd,jones2009development}. Also, there are algorithmic methods for clustering COPD patients based on big data, complex network analysis, and deep learning \cite{burgel2017simple,divo2015chronic,chang2016copd,min2019predictive,mihaicuta2017network}. {However, these techniques have not resulted in high prediction accuracy; the reason is that they only focus on investigating novel machine learning models rather than analyzing the geometric characteristics of the data. Furthermore, big data and Internet-of-Things (IoT) solutions were proven to be effective in COPD management, but such existent engineering systems are merely monitoring physiological signals to provide therapeutic feedback to physicians \cite{ding2012mobile,hardinge2015using}. }
Instead, in this paper, our main objective is to investigate the geometric features of the COPD physiological signals and offer a rigorous alternative to the conventional (spirometry-based) methodology for COPD diagnostics. Two hypotheses underpin the solution we propose:

\begin{enumerate}
\item The physiological signals relevant to COPD (e.g., respiratory rate, oxygen saturation, abdomen breathing effort, etc.) have a multi-fractal nature, and their fractional-order dynamics specifically characterize the COPD pathogenic mechanisms.
\item We can capture the fingerprints of the COPD-related physiological processes with the coupling matrix in our mathematical modeling of the physiological dynamics. (In other words, the coupling matrix $A$ deciphers the interdependencies and correlations between the recorded signals.)
\end{enumerate}
{In this work, we generate two noval COPD physiological signals datasets (WestRo COPD dataset and WestRo Porti COPD dataset) and implement our method by analyzing the relevant physiological signals recorded with an IoMT (Internet of Medical Things) infrastructure. Then, we extract the fractional dynamics signatures specific to the COPD cases and train a deep neural network to predict COPD stages using both fractal dynamic network signatures and expert analysis (see Figure 1). Our method demonstrates a high prediction accuracy of $98.66\%\pm0.45\%$ for the WestRo COPD dataset (4432 medical cases gathered from COPD patients). The $k$-fold ($k=5$) cross-validation and hold-out validation confirm these results (to test the generalization capabilities of our framework, we also utilize the transfer learning mechanism to validate the fractional dynamic deep learning model with an external WestRo Porti COPD dataset where sleep apnea and COPD coexist in many cases). For detailed information about our datasets, see \emph{Methods} section \emph{Data collection}.}

\begin{figure}

\includegraphics[width=\textwidth]{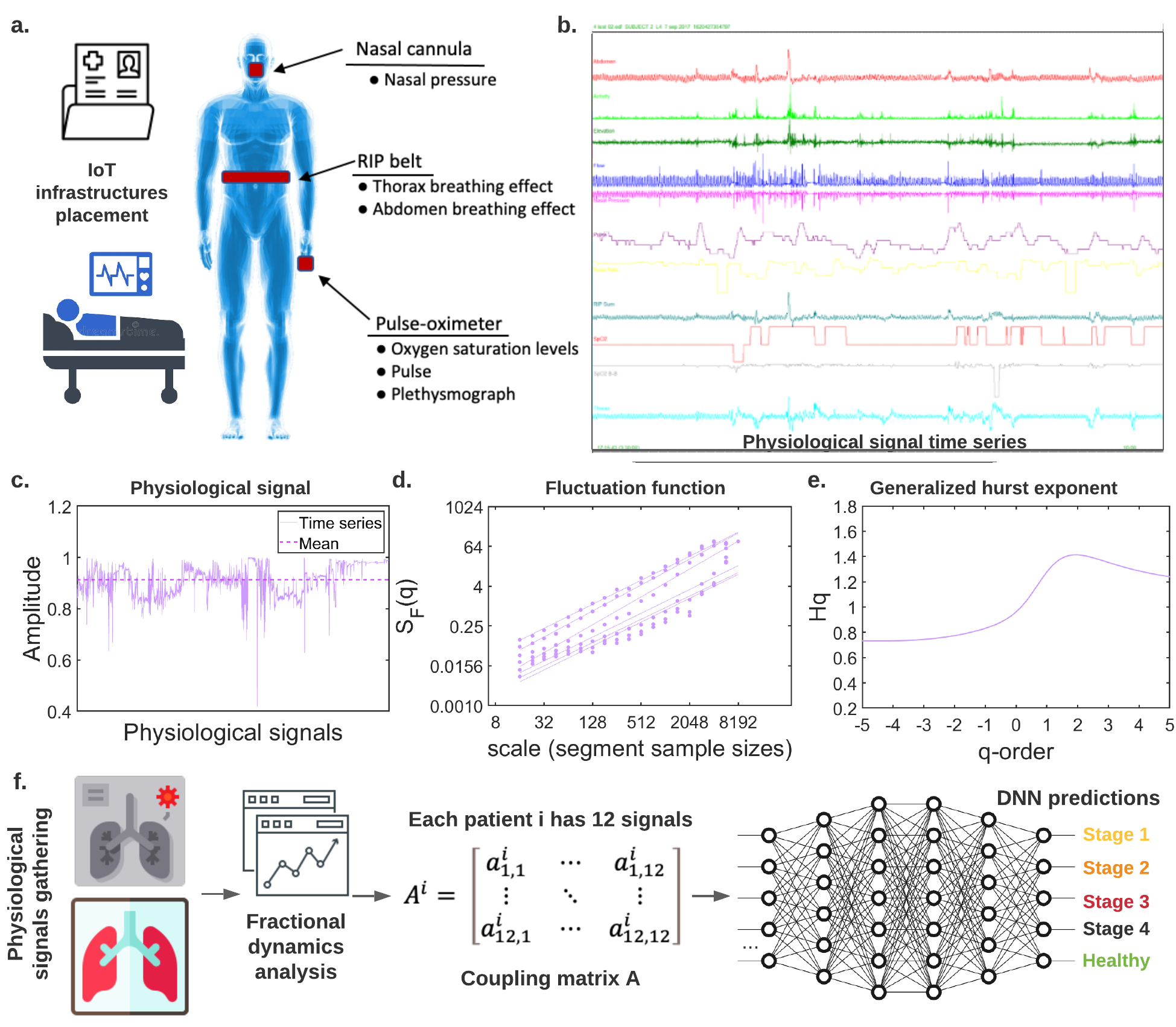}

\caption{\textbf{Overview of the proposed method for COPD stage prediction}: (a) based on the medical observations from the latest research in the field, we identify the physiological signals with relevance in COPD and measure them, (b) we record such physiological signals with a medical sensor network---NOX T3\texttrademark\ portable sleep monitor \cite{cairns2014pilot}. (c) For each signal, we perform (d) fluctuation analysis and (e) long-term memory analysis to investigate the fractional features among medical cases. (f) We employ the analysis of the fractional-order dynamics to extract the signatures of the signals as coupling matrices and fractional-order exponents and use these signatures (along with expert diagnosis) to train a deep neural network that can identify COPD stages.}
\label{fig:fig1}
\end{figure}


\section*{Results}

\subsection*{Physiological signals}

\subsubsection*{Recorded signals}

\noindent
\textbf{WestRo COPD dataset.} In this dataset, each medical case consists of 12 signal records. First, we recorded seven physiological signals from our patients with the Respiratory Inductance Plethysmography--RIP (signals Thorax Breathing Effort and Abdomen Breathing Effort), the wireless pulse-oximeter (signals Oxygen Saturation Levels, SpO2 beat-to-beat mode, Pulse, and Plethysmograph), and the nasal cannula (signal Nasal Pressure). The NOX T3\texttrademark\ portable sleep monitor integrates and synchronizes the RIP, the wireless pulse-oximeter, and the nasal cannula.  (Section \emph{Methods}, subsection \emph{Data collection} provides detailed information). Moreover, the Noxturnal\texttrademark\ software application, which accompanies the NOX T3\texttrademark\, derived five additional signals: RIP Sum (the sum of the abdomen and thorax breathing effort signals), Activity (derived from the X, Y, and Z gravity axes), Position (in degrees, derived from the X, Y, and Z gravity axes, where the supine position is 0 degrees), Flow (derived from the nasal pressure signal), Resp Rate (respirations per minute derived from the RIP Sum signal). \\

\noindent
\textbf{WestRo Porti COPD dataset.} The dataset consists of 6 physiological signals recorded in 13824 medical cases from 534 individuals during 2013-2020. The patients in the WestRo Porti are screened with the Porti SleepDoc 7 potable PSG device by recording 6 physiological (signals Flow, SpO2, Pulse, Pulsewave, Thorax, Abdomen) overnight. The 6 Porti SleepDoc 7 signals correspond, respectively, to the following NOX T3 signals: Flow, Oxygen Saturation Levels, Pulse, Plethysmograph, Thorax Breathing Effort, and Abdomen Breathing Effort. The reason for involving this dataset in this study is that: (1) we want to involve an external dataset to test the validation of our model; (2) we want to test the robustness of our prediction and diagnosis approach where the medical signals records in this dataset interfere with another disease (sleep apnea). 

\subsubsection*{Fractal properties of physiological signals}

To verify our first hypothesis, in this section, we show the fractal features in non-derived signals (Thorax, Oxygen Saturation, Pulse, Plethysmograph, Nasal Pressure, and Abdomen) between healthy persons (stage 0) and critical COPD patients (stage 4) in our WestRo COPD dataset. As shown in \cite{kantelhardt2002multifractal, peng1994mosaic}, Detrended Fluctuation Analysis (DFA) is an effective method to determine the statistical self-affinity of a stationary fractional Gaussian noise (fGn) type or non-stationary fractional Brownian motion (fBm) type signal. In order to mine the physiological complexity and account for its nonstationarity, we perform a comprehensive multifractal detrended fluctuation analysis (MF-DFA) of the collected data. 

\begin{figure}[h!]
\centering
\includegraphics[width=\textwidth]{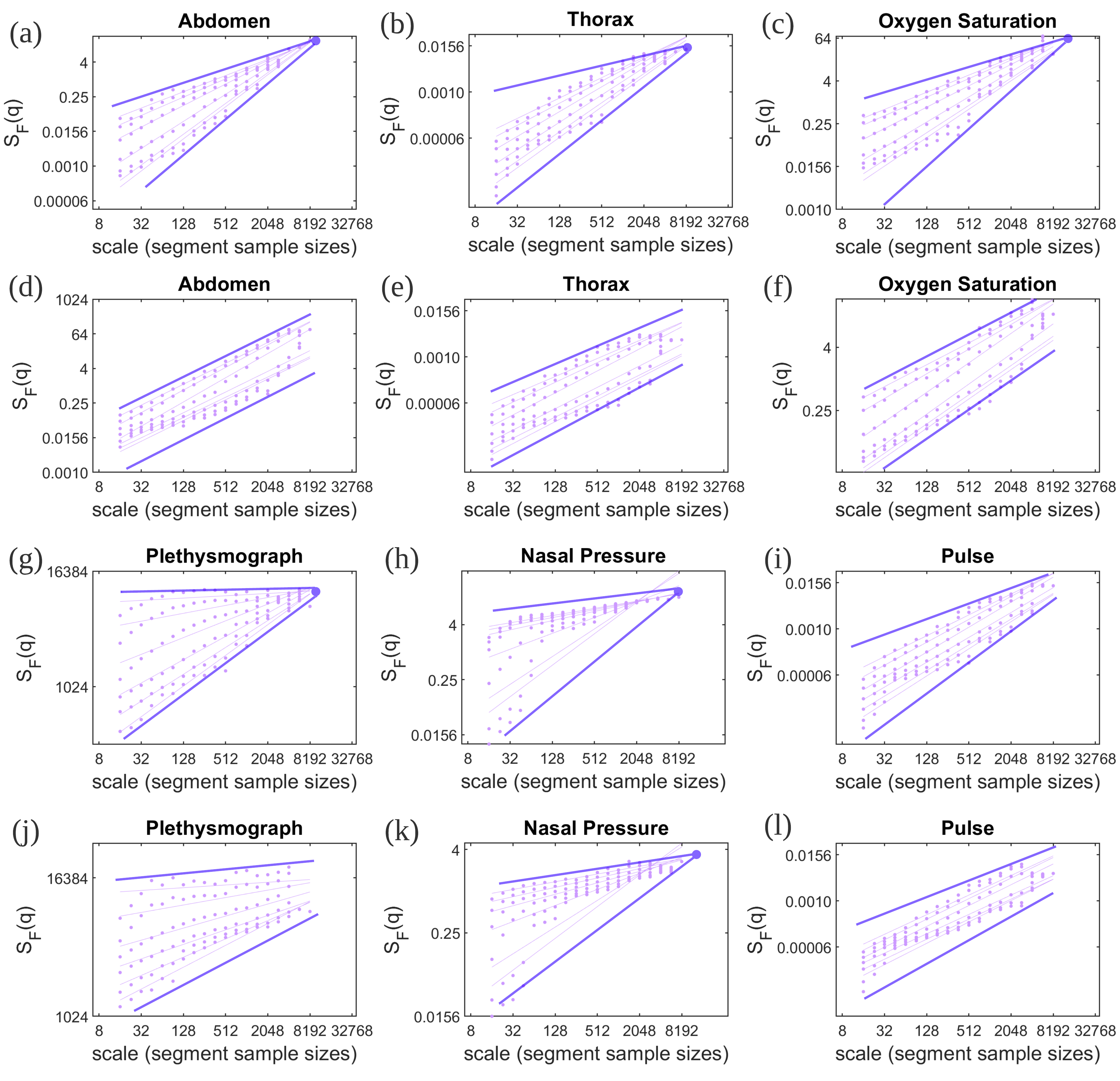} 
\caption{{\textbf{The geometry of fluctuation profiles for the COPD-relevant physiological signals recorded from a normal abdomen and a stage 4 COPD abdomen.} We calculate the scaling functions from 6 non-derived physiological signals: Abdomen, Thorax, Oxygen Saturation, Plethysmograph, Nasal Pressure, and Pulse, where the exponents are $q\in$ [-5, 5]. Panels (a-c) and (g-i) with signals recorded from a healthy person, (d-f), and (j-l) with signals recorded from a representative stage 4 COPD patient. The resulting points (pink nodes) of the multifractal scaling function with power-law scaling will converge to a focus point (dark purple nodes) at the largest scale, $L$.}}
\label{fig:hyp_test}
\end{figure}

To analyze the fractional dynamic characteristics of the COPD physiological processes, we calculate the scaling (fluctuation) functions of the non-derived signals (Thorax, Oxygen Saturation, Pulse, Plethysmograph, Nasal Pressure, and Abdomen) in healthy people (stage 0) and very severe COPD patients (stage 4) (for detailed information about \emph{MF-DFA} and the \emph{scaling function}, see section \emph{Methods}, subsection \emph{Multifractal detrended fluctuation analysis}). In Figure \ref{fig:hyp_test}, panels (a-c) and (g-i), respectively, show the scaling functions calculated from Abdomen, Pulse, Plethysmograph, Thorax, Nasal pressure, and Oxygen saturation signals generated for a healthy person and panels (d-f) and (j-l) illustrate the scaling functions for the same signals from a very severe COPD patient (stage 4). We set the $q$ values as $q\in \left\{-5, -3, -1, 1, 3, 5\right\}$. In Figures \ref{fig:hyp_test} (a-c) and (g-i), we find that the scaling functions under different $q$ values will converge to a focus point (the dark purple nodes in each panel), except the Pulse signal. The focus $S(v,L)$ can be measured as the scaling function's fundamental property (for details on \emph{force point}, see section \emph{Methods}, subsection \emph{Multifractal detrended fluctuation analysis}). This way, if a signal's scaling function has a focus point, it is a multifractal signal. {The pink lines in Figure \ref{fig:hyp_test} are the lines best-fitted to the scaling function's observed data and the pink dots represents different segmented sample sizes. To be more specific, we show the scaling functions of physiological signals extracted from all stage 4 patients and the healthy people (stage 0) in our dataset (where $q\in[-5, 5]$) in Supplementary material (for detailed information, see \emph{Supplementary material}'s section \emph{Hurst exponents of physiological signals}). Hence, from Figures \ref{fig:hyp_test} (a-c, g-i), the non-derived signals---except the Pulse signal---in healthy persons have multifractal features. Conversely, in Figure \ref{fig:hyp_test} (d-f, j-l), these scaling functions do not have focus points within the scale (except the Nasal pressure signal), which may suggest that such signals---recorded from severe COPD patients---do not have multifractal features.} 

\begin{figure}[ht]
\centering
\includegraphics[width=\textwidth]{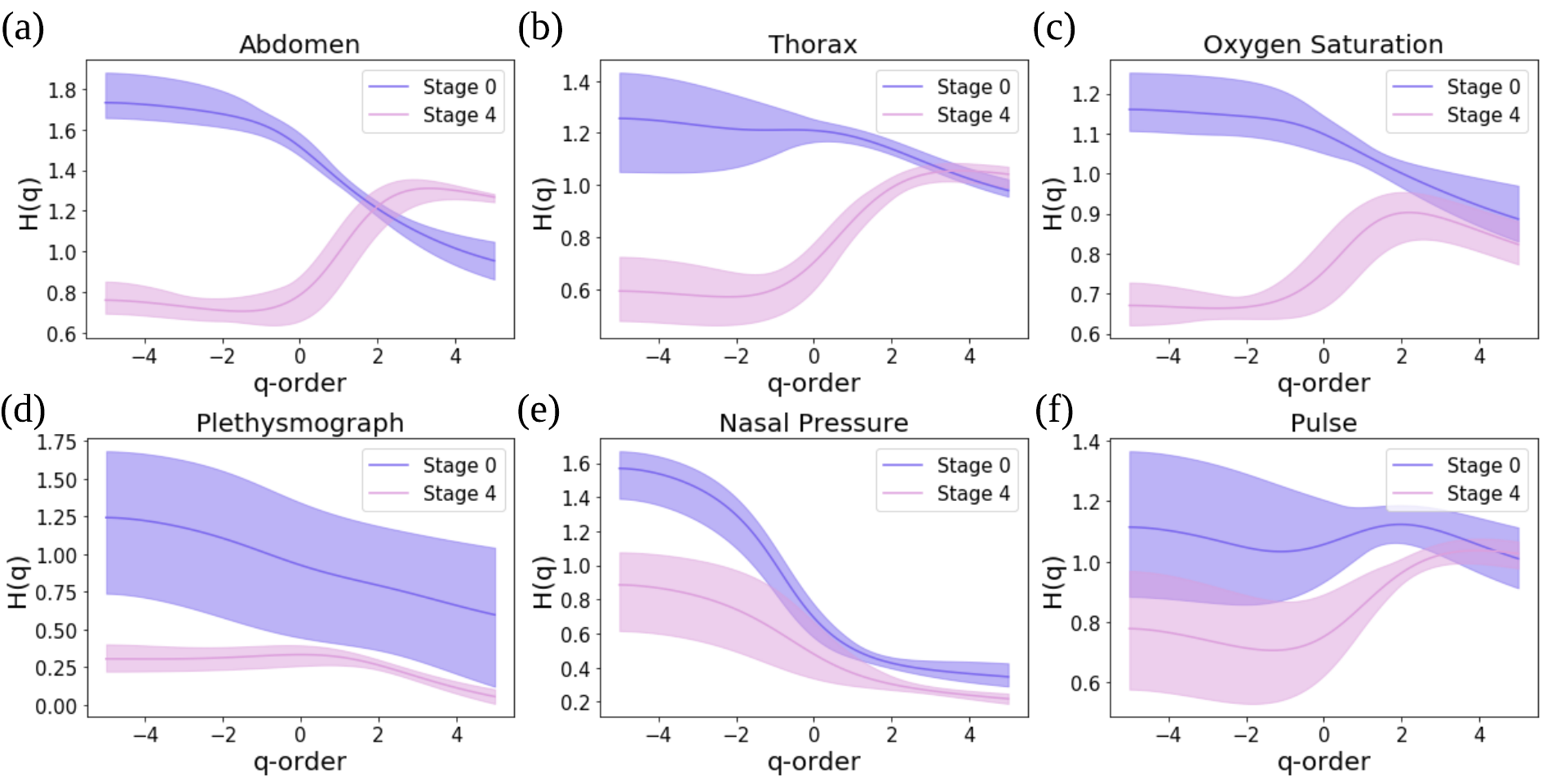}
\caption{\textbf{Multifractal analysis of 6 physiological signals from healthy people (stage 0) and stage 4 COPD patients with 95\% confidence interval:} Generalized Hurst exponent $H(q)$ as a function of $q$-th order moments (where $q$ values are discretely extracted from $-5$ to $5$) for physiological signals (Abdomen (a), Thorax (b), Oxygen Saturation (c), Plethysmograph (d), Nasal Pressure (e), and Thorax (f)) extracted from healthy people (stage 0) and severe COPD patients (stage 4).}
\label{fig:hq}
\end{figure}

Figure \ref{fig:hq} presents the $H(q)$ comparisons between physiological signals (Abdomen (a), Thorax (b), Oxygen Saturation (c), Plethysmograph (d), Nasal Pressure (e), and Thorax (f)) extracted from healthy people and patients with 95\% confidence interval, where $H(q)$ is the Hurst exponent and represents the set of associated slopes of the pink lines in Figure \ref{fig:hyp_test}. 
The Hurst exponent measures the long-term memory of time series, and larger Hurst exponent values (i.e., $H(q)>0.5$) represent slower evolving variations (more persistent structure) in mono- and multi-fractal time series\cite{ihlen2012introduction}.
$H(q)$ confidence intervals among healthy people (stage 0) and COPD patients (stage 4) in Figure \ref{fig:hq} have different functional trends/distributions, which shows that the physiological signals extracted from healthy people and severe COPD patients possess different multi-fractional properties. (The non-linear decrease distribution of the $H(q)$ function illustrates that the fitted lines of scaling functions under different $q$ values will converge to a force point). In Figure \ref{fig:hq}, all signals exhibit long-range dependency properties ($H(q)>0.5$) except the Plethysmograph signals extracted from COPD patients; furthermore, all the signals extracted from healthy people present a slower evolving variation behavior than the signals gathered from COPD patients. Consequently, the overarching conclusion of Figure \ref{fig:hyp_test} and Figure \ref{fig:hq} is that the physiological signals with relevance for COPD have different fractional dynamics characteristics in healthy people---on the one hand---and severe COPD patients---on the other hand.

\begin{figure}[h]
\centering
\includegraphics[width=\textwidth]{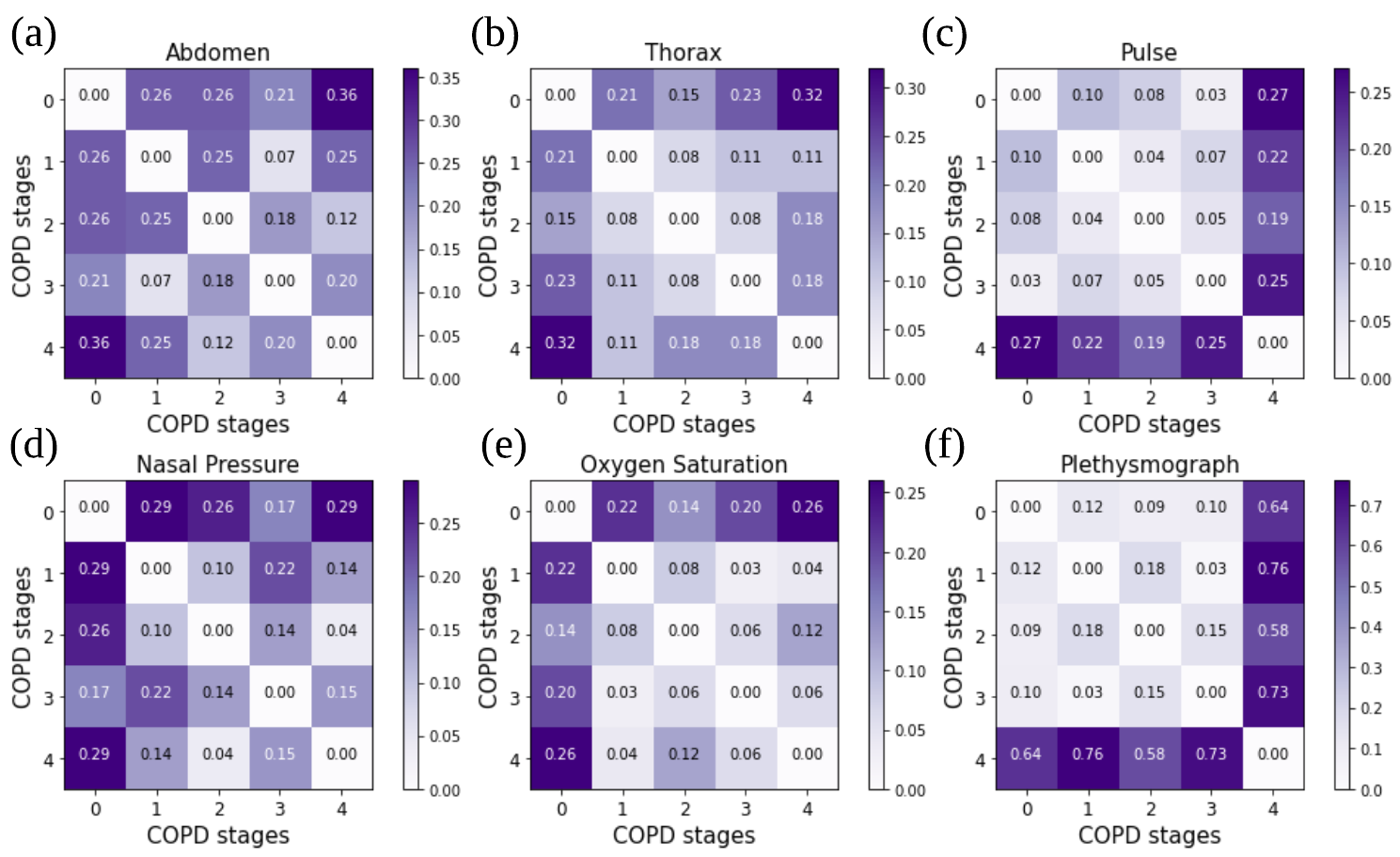} 
\caption{{\textbf{Comparison of Wasserstein distance between the distributions of mean $H(q)$ curves across different COPD stages (i.e., $H(q)$ curves in Figure~\ref{fig:hq}, Figure S2, and Figure S3).}  Comparison in terms of Wasserstein distance between the $H(q)$ distribution function for Abdomen (a), Thorax (b), Pulse (c),  Nasal Pressure (d), Oxygen Saturation (e), and Plethysmograph (f) signals recorded from patients across all the COPD stages. }}
\label{fig:w_dis}
\end{figure}

{To analyze the Hurst exponent distribution curves across different COPD stages, we calculate the Wasserstein distance between each $H(q)$ mean value curve in every physiological signal extracted from patients with different stages. (For detailed results about $H(q)$ curves for all COPD stages, see \emph{Supplementary material}'s section \emph{Hurst exponents of physiological signals}.) The Wasserstein distance is a metric that measures differences between distributions~\cite{olkin1982distance}. In Figure~\ref{fig:w_dis}, we show that, for each physiological signal, the signals' $H(q)$ function extracted from stage 0 patients have the largest Wasserstein distance from stage 4 patients (except the Plethysmograph signal). In contrast, the $H(q)$ functions of signals extracted from patients in moderate to severe COPD stages (i.e., 1, 2, and 3) have much smaller Wasserstein distances between each other (except the Abdomen signal). The results presented in Figure~\ref{fig:w_dis} represent evidence that reinforces the medical observation stating that it is hard to distinguish early COPD stages. As such, it makes sense to analyze the spatial coupling between these physiological processes (signals) across time. These spatial coupling matrices $A$ contain the different fractional features across different signal samples and can help us classify the signals recorded from suspected patients into different COPD stages.}

The general conclusion in Figures \ref{fig:hyp_test}, \ref{fig:hq}, and \ref{fig:w_dis} is that the physiological signals have multifractal properties in healthy individuals but do not have such features in COPD patients. (This dichotomy is less evident for the Pulse signal, even if the tendency towards multifractality in healthy individuals is still present.) The notable exception is the Nasal Pressure signal, which has multifractality in both healthy and COPD individuals. This observation suggests that COPD mostly affects the physiology of muscles involved in or supporting breathing (Thorax Breathing Effort and Abdomen Breathing Effort) \cite{jaitovich2018skeletal,mathur2014structural} and the circulatory system's physiology (reflected in Oxygen Saturation Levels and Plethysmograph)\cite{elbehairy2015pulmonary}. Conversely, the upper respiratory tract's physiological dynamics do not appear to be affected even by severe and very severe COPD stages. 

\subsection*{Fractional dynamics modeling of COPD relevant physiological signals subject to unknown perturbations}

The dynamics of complex biological systems possess long-range memory (LRM), non-local, and fractal characteristics mathematically modeled through fractional calculus concepts. For instance, several recent studies have demonstrated that stem cell division times~\cite{11_bogdan2014heterogeneous}, blood glucose dynamics~\cite{gupta2018dealing}, heart rate variability~\cite{ivanov1996,ivanov1999} and brain-muscle interdependence activity~\cite{13_xue2016spatio} are either fitted by power-law distributions or obey fractal statistics ~\cite{12_ghorbani2013cyber,14_xue2016minimum,15_xue2017constructing,yin2020discovering, gupta2021non}. The long short-term memory (LSTM) architecture is one of the most widely used deep learning approaches to analyze biological signals and perform prediction or classification. However, LSTM cannot fully represent the long memory effect in the input, nor can it generate long memory sequences from unknown noise inputs\cite{greaves2019statistical,cheng2016long}. Thus, when considering the very long-time series with long-range memory, LSTM cannot predict nor classify them with high accuracy. Indeed, in our study, the length of each physiological signal has more than 72000 data points. We aim to capture both short-range and long-range memory characteristics of various physiological processes and---at the same time---investigate the very long COPD signals with high accuracy; therefore, we adopt the generalized mathematical modeling of the physiological dynamics,
\begin{eqnarray}
\Delta^{\alpha}x[k+1] &=& Ax[k] + Bu[k] \nonumber\\
y[k] &=& Cx[k], 
\label{eqn:fracLlinModel}
\end{eqnarray}
\noindent where $x\in\mathbb{R}^{n}$ is the state of the biological system, $u \in \mathbb{R}^{p}$ is the unknown input and $y \in \mathbb{R}^{n}$ is the output vector\cite{gupta2018dealing}. The main benefits of this generalized mathematical representation are three-fold:  
\begin{enumerate}
\item The model allows for capturing the intrinsic short-range memory and long-range memory of each physiological signal through either an integer or fractional order derivative. To connect the mathematical description with the discrete nature of measurements, the differential operator $\Delta$ is used as the discrete version of the derivative; for example, $\Delta^{1}x[k] = x[k] - x[k-1]$. A differential order of $1$ has only one-step memory, and hence the classic linear-time invariant models are retrieved as particular cases of the adopted mathematical model. However, when the differential order is $1$, the model cannot capture the long-range memory property of several physiological signals. Furthermore, we write the expansion of the fractional derivative and discretization \cite{andrzej} for any $i\textsuperscript{th}$ state $(1\leq i\leq n)$ as 
\begin{equation}
\Delta^{\alpha_{i}}x_{i}[k] = \sum\limits_{j=0}^{k}\psi(\alpha_{i},j) x_{i}[k-j],
\label{eqn:fracExpan}
\end{equation}
\noindent where $\alpha_{i}$ is the fractional order corresponding to the $i\textsuperscript{th}$ state and $\psi(\alpha_{i},j) = \frac{\Gamma(j-\alpha_{i})}{\Gamma(-\alpha_{i})\Gamma(j+1)}$ with $\Gamma(.)$ denoting the gamma function. Equation (\ref{eqn:fracExpan}) shows that the fractional-order derivative framework provides a mathematical approach to capture the long-range memory by including all $x_{i}[k-j]$ terms.
\item Our modeling approach describes the system dynamics through a matrix tuple $(\alpha, A, B, C)$ of appropriate dimensions. The coupling matrix $A$ represents the spatial coupling between the physiological processes across time, while the input coupling matrix $B$ determines how the inputs affect these processes. We assume that the input size is always strictly smaller than the state vector's size, i.e., $p < n$. The coupling matrix $A$ plays an essential role in deciphering the correlations between the recorded physiological signals. These correlations (entries of $A$) can indicate different physical conditions. For instance, when probing the brain electrical activity (through electroencephalogram (EEG) signals), the correlations can help at differentiating among various imaginary motor tasks \cite{gaurav2018BCI}. Moreover, as described in this work, we can exploit these correlations to differentiate among pathophysiological states---such as degrees of disease progression---using physiological signals analysis. A key challenge is the estimation accuracy of these correlations ($A$ matrix), notably for partially observed data. We have taken care of such limitations by using the concept of unknown unknowns introduced in reference \cite{gupta2018dealing}. 

\item Since we may have only partial observability of the complex biological systems, we take care of the unknown stimuli (excitations that may occur from other unobserved processes but cannot be probed); as such, we include in the model the vector variable $u$ and study its impact on the recorded dynamics. In essence, we refer to this mathematical model as a multi-dimensional fractional-order linear dynamical model with unknown stimuli. 
 The model parameters are estimated using an Expectation-Maximization (EM) based algorithm described in reference \cite{gupta2018dealing}, to overcome the lack of perfect observability and deal with possibly small and corrupted measurements. Reference \cite{gupta2018dealing} proves that the algorithm is convergent and shows that it reduces modeling errors.
\end{enumerate}

\subsection*{Fractional Dynamics Deep Learning Prediction of COPD stages}

After extracting the signals' features (short-range and long-range memory) with the fractional dynamic mathematical model, we utilize these features (i.e., coupling matrices $A$) to train a deep neural network to predict patients' COPD stage. Deep learning is a machine learning approach that efficiently combines feature extraction and classification and it is a valuable tool for medical diagnosis (i.e., it can logically explain a patient's symptoms \cite{gunvcar2018application, yin2023anatomically}). We develop the fractional dynamics deep learning model (FDDLM) presented in this section to predict the COPD stages for our WestRo COPD dataset consisting of 4432 medical cases from patients in Pulmonology Clinics from Western Romania. We evaluate these cases and FDDLM by $k$-fold cross-validation and hold-out validation. $K$-fold cross-validation is a resampling procedure used to estimate how accurately a machine learning model will perform in practice. In $k$-fold cross-validation ($k=5$), we randomly shuffle the input dataset and split it into 5 disjoint subsets. We select each subset as the test set ($20\%$) and combine the remaining subsets as the training set ($80\%$). In hold-out validation, we hold one institution out at a time. I.e., we hold out data from one institution as a test set, and the remaining data from the other three institutions are used to train the models. The main steps of our approach are: 

\begin{enumerate}
    \item We construct a COPD stage-predicting FDDLM, and calculate coupling matrix signatures ($A$) of relevant physiological signals (such as Thorax Breathing Effort or Abdomen Breathing Effort, etc.) to be used as the training data.
    \item We train FDDLM with our training set to recognize the COPD level based on signal signatures. 
    \item We test FDDLM with our test set and predict patients' COPD stage. 
 \end{enumerate}
 
\begin{figure}
\centering
\includegraphics[width=\textwidth]{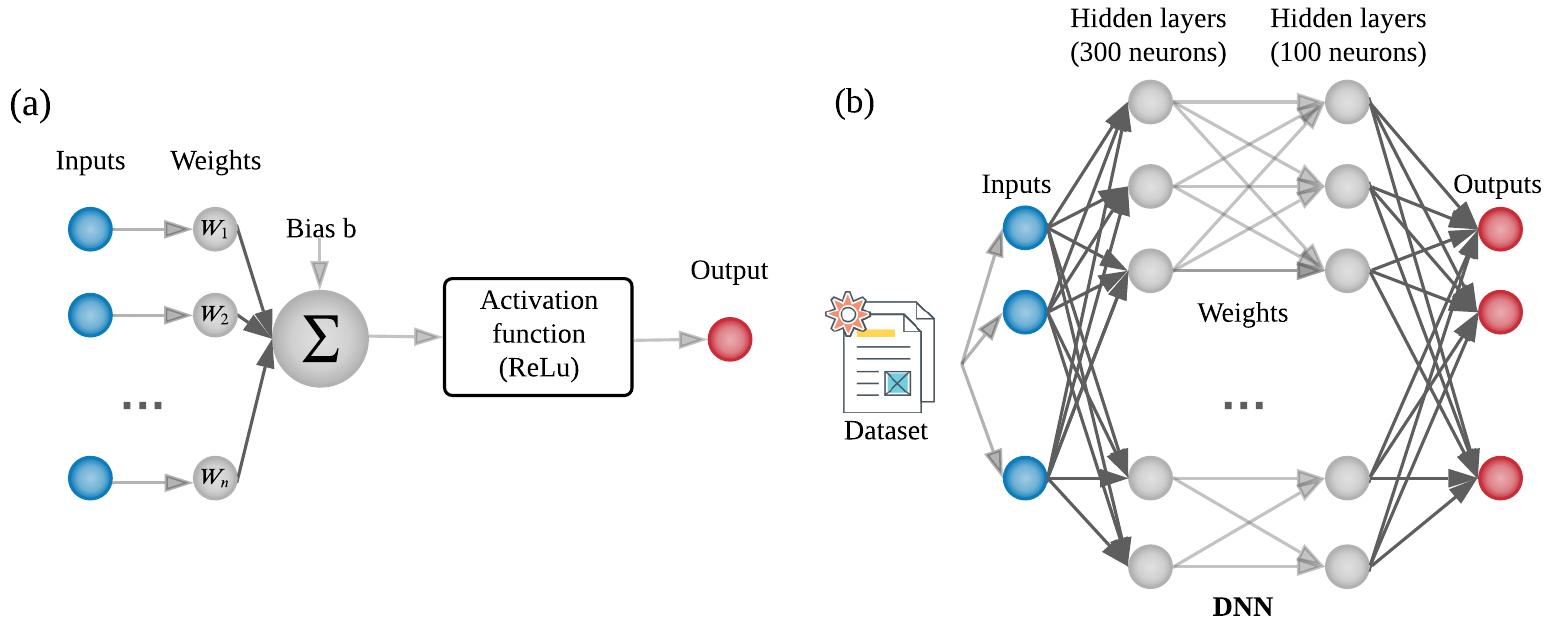}
\caption{ \textbf{Network formulation for FDDLM:} (a) Basic structure of an artificial neuron model; (b) Overview of the neural network model we trained in FDDLM to identify COPD stages.}
\label{fig:dnn}
\end{figure}

 Our FDDLM uses a feedforward deep neural network architecture \cite{lecun2015deep} with four layers: one input layer, two hidden layers, and one output layer. We present our model's structure in Figure \ref{fig:dnn}. The input layer takes the input (i.e., signal signatures in the coupling matrix $A$) and pass it to hidden layers. From layer to layer, neurons compute the sum of the weighted inputs from the previous layer and settle the results through a nonlinear function (activation function) (see Figure \ref{fig:dnn} (a)). In our FDDLM, hidden layers' activation functions are \emph{rectified linear unit} ($ReLU$)\cite{ian2016deep}. The $ReLU$ activation function returns the input value when the input is $\leq0$ (otherwise, it returns 0), i.e., $g(z)=max(0,z)$ for input value $z$. The output layer's activation function is \emph{softmax} ($S$), which normalizes the input values into a probability distribution. We utilized the \emph{rmsprop} optimizer (with the default learning rate of 0.001) and the \emph{categorical cross entropy} loss function. To avoid potential overfitting, we insert \emph{dropout} function (with 0.8 keep rate) after each hidden layer to randomly select and ignore 20\% of neurons.

\begin{figure}[]
\centering
\includegraphics[width=1\textwidth]{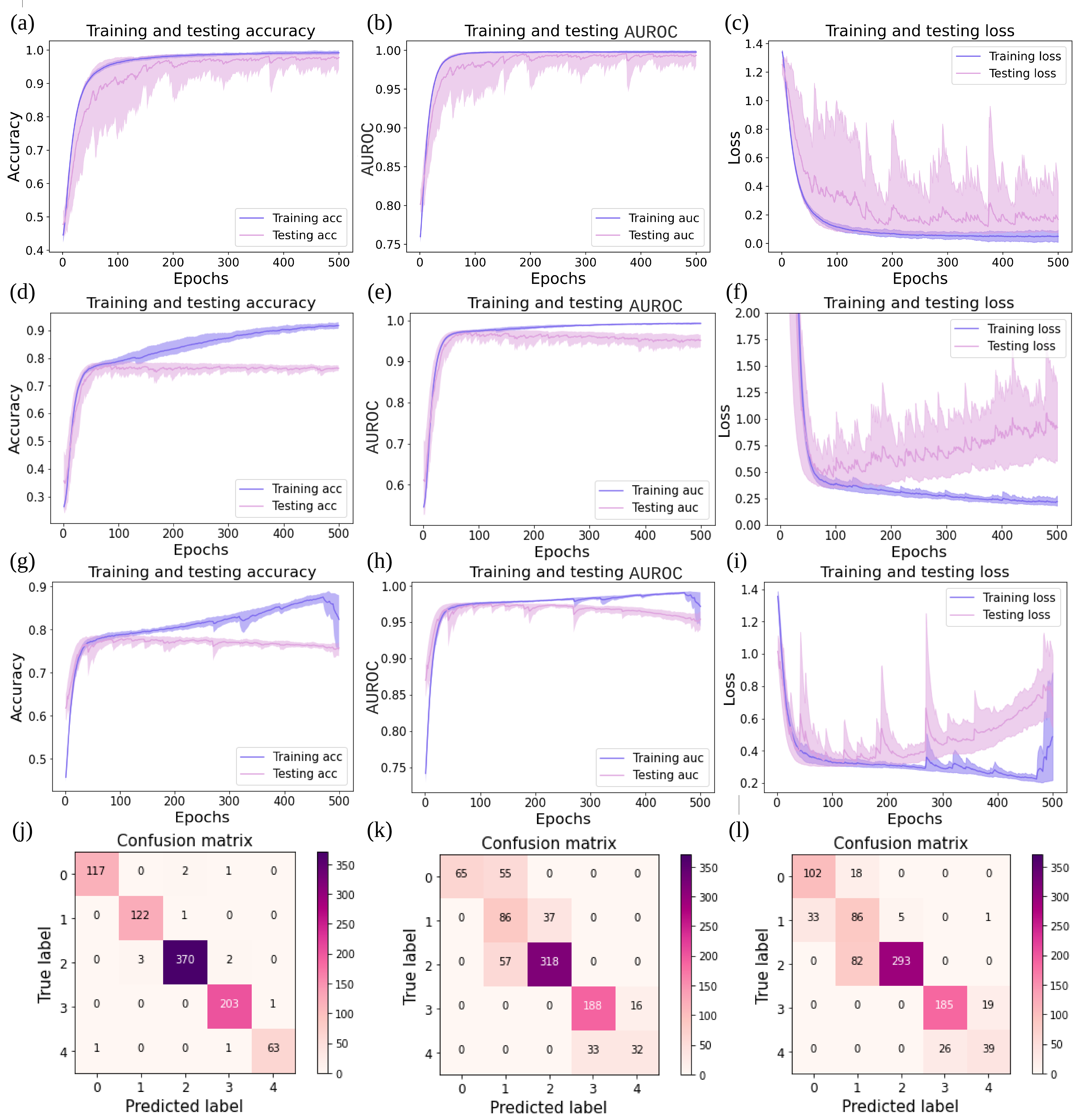}
\caption{\textbf{Training and testing result comparisons (accuracy (a,d,g), AUROC (b,e,h), and loss (c,f,i)) of different deep learning models for the $k$-fold cross-validation.} Training/testing accuracy (a), AUROC (b), and loss (c) for our FDDLM, where the training processes use signal signatures extracted with the fractional dynamic mathematical model. Training/testing accuracy, AUROC, and loss for the Vanilla DNN model (d-f) and the LSTM model (g-i), where the training processes use the physiological signals recorded with portable sleep monitors (raw data). Both Vanilla DNN and LSTM models share similar network structures and computation parameters with our FDDLM (Vanilla DNN has the same network structure as our model, except the input size). We obtain these results with the $k$-fold cross-validation ($k=5$). We also show the confusion matrices for test set across different models: FDDLM in panel (j), Vanilla DNN in (k), and LSTM in (l).}
\label{fig:acc}
\end{figure}
\subsubsection*{$K$-fold cross-validation results}
\begin{table}
\renewcommand\arraystretch{1.2}
\centering
\caption{\label{tab:1} The COPD stage predicting results for test set with our Fractional Dynamics Deep Learning Model (FDDLM).}
\resizebox{10cm}{!}{%
\begin{tabular}{cccccc}
\toprule
COPD & Stage 0 & Stage 1 & Stage 2 & Stage 3 & Stage 4\\ \hline
Sensitivity    & 97.50\% & 99.19\% & 98.66\% & 99.50\% & 96.92\% \\
Specificity    & 99.87\% & 99.61\%  & 99.41\%  & 99.31\% & 99.88\% \\ 
Precision    & 99.15\% & 97.60\%    & 99.20\%  & 98.06\% & 98.44\%\\ 
\bottomrule
\end{tabular}%
}
\end{table}

We process all physiological signals with the fractional dynamical model \cite{gupta2018dealing}; then, we feed the signal signatures from the coupling matrix $A$ to FDDLM. We implement the neural network model in Python with Keras package and executed it on a computer with the Intel Core i7 2.2GHz processor and 16GB RAM.

We evaluate our results based on accuracy, sensitivity, loss, precision, specificity, and area under the receiver operating characteristic curve (AUROC). Our estimation of all results---generated from different models---uses the $k$-fold cross-validation method (with $k=5$). Figure \ref{fig:acc} presents our model's accuracy, AUROC, and loss curves on training and test sets. We also evaluate our model by comparing it with the Vanilla DNN and LSTM models trained on physiological signals (i.e., raw data) extracted from sleep monitors for reference. The Vanilla DNN and LSTM models have the same hyper-parameter setting as ours, including the optimizer, dropout configuration, loss function, activation functions, except the input layer size. We aim to choose the classic deep learning models with similar structures as baselines to investigate whether the fractional features (coupling matrix $A$) are easier for models to classify than the raw data. Figures \ref{fig:acc} (a-c) show the training and test results observed from FDDLM. We observe that training and test accuracies and AUROC increase, while loss decreases during training.



Figure~\ref{fig:acc} (d-f) and (g-i) present the training and test accuracy, AUROC, and loss results of Vanilla DNN model and LSTM model trained with physiological signals (raw data). {The training and test results obtained from both Vanilla DNN model and LSTM model display overfitting, as test accuracies decrease (test loss increase) while training accuracies increase (training loss decrease). Thus, to deal with overfitting, we involve the early-stopping technique~\cite{raskutti2014early} in choosing the best-performing Vanilla DNN and LSTM. Compared to our FDDLM, the best-performing Vanilla DNN and LSTM result in much lower accuracies: $77.72\%\pm0.688\%$ and $78.54\%\pm1.200\%$, respectively, while our FDDLM achieves $98.66\%\pm0.447\%$.}

Figures \ref{fig:acc} (j-l) show the confusion matrix examples of FDDLM, Vanilla DNN, and LSTM, respectively. The confusion matrices present the prediction results of the test set. (We construct the test set using the last $20\%$ of data in the WestRo COPD dataset and train models with the first $80\%$ of data.)

The results point out that the FDDLM only misclassified $1.35\%$ of the test sets in terms of individual COPD stages. Instead, Vanilla DNN and LSTM models misclassified $24.21\%$ and $21.41\%$ of the test sets, respectively. {(We also investigated the possibility of using the convolutional neural network (CNN) model to characterize the physiological signal dynamics obtained from sleep monitors (raw data) and compare it with our FDDLM. The CNN model misclassified $63.87\%$ of the test sets with $k$-fold cross-validation. For detailed information about CNN model and results, see section \emph{Methods}, subsection \emph{Neural network architecture for the WestRo COPD dataset} and \emph{Supplementary material}'s subsection \emph{Training and testing results for CNN}.)} Table \ref{tab:1} presents the precision, sensitivity, and specificity of our model's predicting results; we find that all these results exhibit a substantial accuracy except the sensitivity of stage 4, which is 96.92\%. In conclusion, our FDDLM predicts patients' COPD stages with a much higher accuracy than Vanilla DNN and LSTM models trained with physiological signals (raw data)---without overfitting---and represents an effective alternative to the spirometry-based diagnostic. (We performed the $K$-fold analysis of our model's accuracy both on a per-recording and a per-patient basis and obtained very similar results; see the Supplementary Information, section \emph{Per-patient based $K$-fold analysis}.)

\begin{figure}[h!]
\centering
\includegraphics[width=1\textwidth]{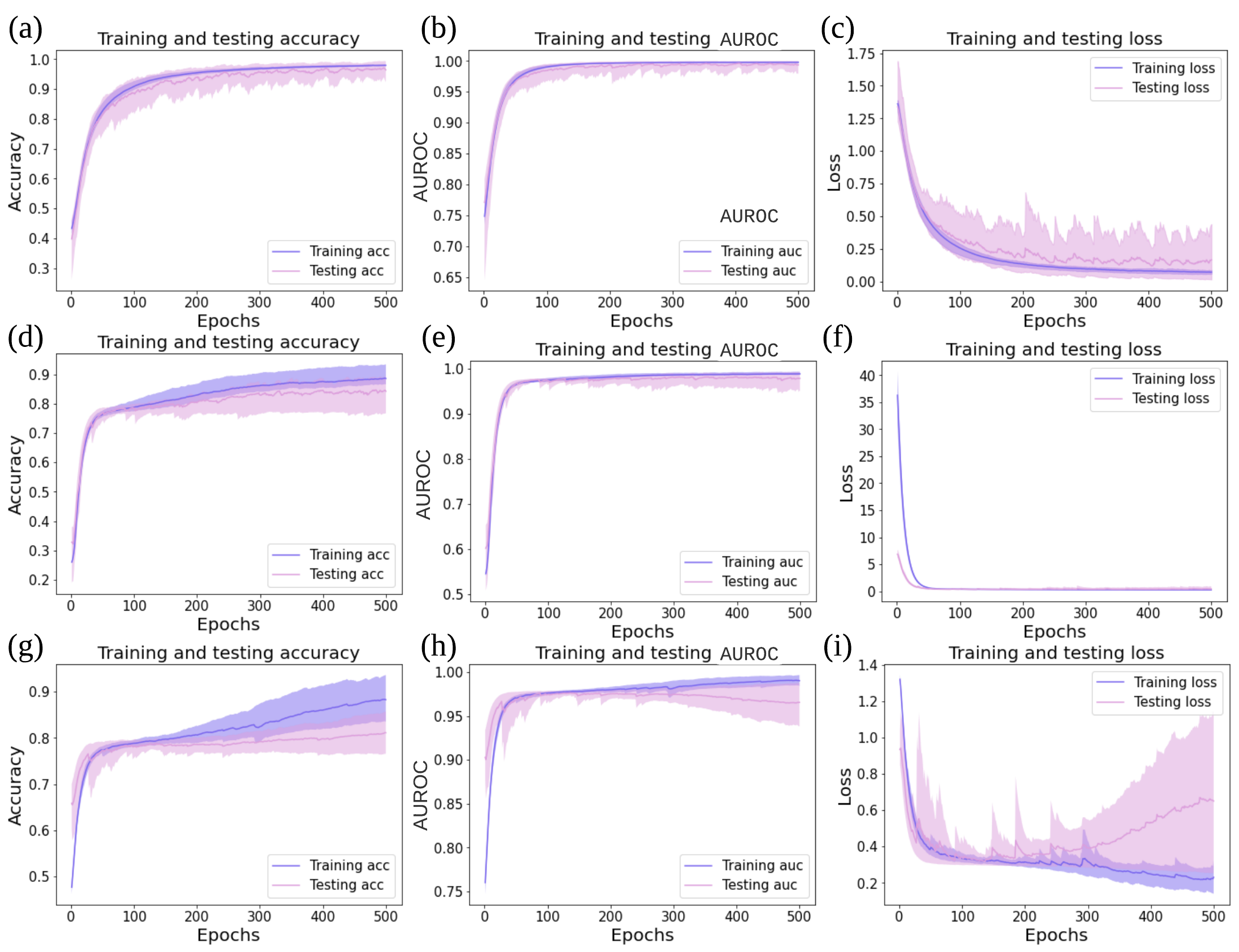}
\caption{\textbf{Training and testing result comparisons of different deep learning models for the hold-out validation.} The training/testing accuracy (a), AUROC (b), and loss (c) for our FDDLM, where the training processes use signal signatures extracted with the fractional dynamic mathematical model. The training/test accuracy (d) and (g), AUROC (e) and (h), and loss (f) and (i), for Vanilla DNN model and LSTM model, where the training processes use the physiological signals recorded with portable sleep monitors. 
Both Vanilla DNN and LSTM models share similar network structures with FDDLM (i.e., same neural network structure but different the input size). We obtain these results by holding out data from every single institution as the test set.}
\label{fig:holdout_curve}
\end{figure}

\begin{figure}[h!]
\centering
\includegraphics[width=1\textwidth]{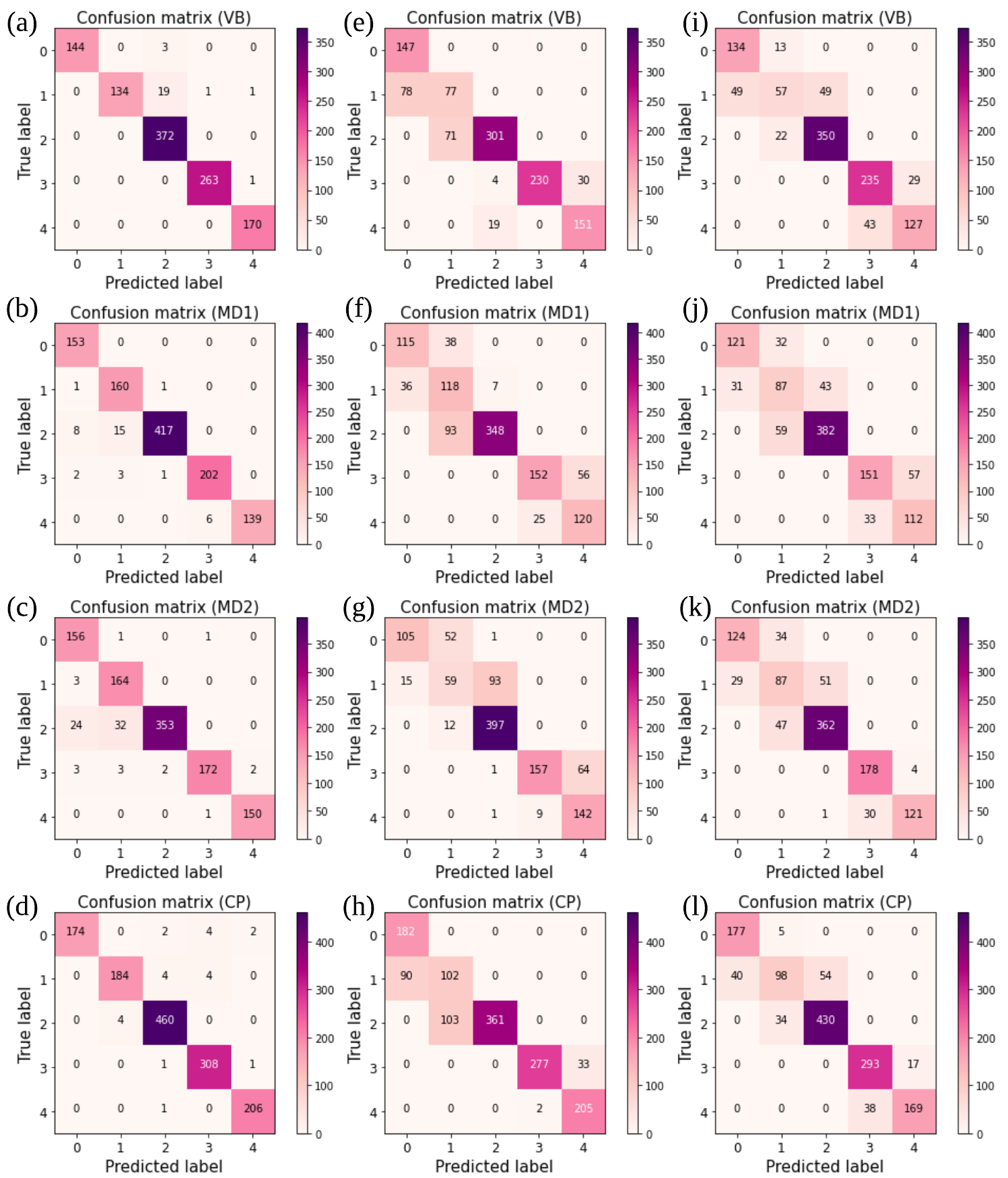}
\caption{\textbf{The comparison of confusion matrices resulted from different deep learning models: fractional dynamics (a-d), Vanilla DNN (e-h), and LSTM (i-l).} We built the test sets by holding out data gathered from one institution (i.e., VB, MD1, MD2, and CP) at a time. The matrix representations clearly show that our model outperforms both Vanilla DNN and LSTM---in all experiments and for all labels representing COPD stages---in terms of prediction errors.}
\label{fig:holdout_cm}
\end{figure}

\subsubsection*{Hold-out validation}

The COPD dataset consists of physiological signals recorded from consecutive patients from four Pulmonology Clinics in Western Romania (Victor Babe\c{s} Hospital -- VB, Medicover 1 -- MD1, Medicover 2 -- MD2, and Cardio Prevent -- CP clinics). To validate our FDDLM, we hold out all data extracted from a single institution as test set and train models on data recorded from the other three institutions. Following experimental setup from the previous section, we use Vanilla DNN and LSTM models as baselines with hyper-parameters similar to our FDDLM. Figure \ref{fig:holdout_curve} shows the results of FDDLM, Vanilla DNN, and LSTM, in terms of accuracy, AUROC, and loss curves of training and test sets. We train Vanilla DNN and LSTM models on physiological signals (i.e., raw data). Conversely, we train our FDDLM on the fractional signatures extracted from the raw data.

Figures \ref{fig:holdout_curve} presents the training and test results (accuracy, AUROC, and loss) generated by FDDLM (a-c), Vanilla DNN model (d-f), and LSTM (g-i) model, respectively. Figure \ref{fig:holdout_curve} (a) shows that the accuracy of FDDLM increases in training without overfitting. {Conversely, the training and testing of Vanilla DNN and LSTM models clearly indicate lower values of the accuracy ($80.73\%\pm3.46\%$ and $80.83\%\pm3.67\%$, respectively) in Figure \ref{fig:holdout_curve} (d) and (g), while FDDLM achieves $95.88\%\pm1.76\%$.}

Of note, we observe that the test accuracy under hold-out validation (95.88\%) is lower than the accuracy obtained under $k$-fold cross-validation (for more detailed error analysis, we present the visualization of extracted features (embeddings) in the last hidden layer of FDDLM across $k$-fold and hold-out validation in the \emph{Supplementary materials} subsection.) The reason for performance degradation in hold-out is that the data recorded from each medical institution are imbalanced. The Victor Babes (VB) and Cardio Prevent (CP) are two large clinics, and COPD patients are more willing to get diagnosis or medical treatment in large units or hospitals rather than small clinics, especially for severe and very severe COPD patients. Thus, the signals gathered from VB and CP are more comprehensive than the Medicover 1's (MD1) and Medicover 2's (MD2). In hold-out validation, although we balance the data across different institutions using over-sampling and under-sampling approaches, the remaining imbalance in data collection is still the leading cause of the prediction accuracy drop in the hold-out section.

Figure \ref{fig:holdout_cm} (a, d, g, j), (b, e, h, k) and (c, f, i, l) show the confusion matrices (we only present the test set prediction results) for FDDLM, Vanilla DNN, and LSTM models by holding out each institution as the test set, respectively. Results shown in Figure \ref{fig:holdout_cm} prove that our model outperforms the baselines in prediction accuracy across the entire hold-out validation process. Especially for the early-stage detection (i.e., stages 1 and 2), our model achieves a higher accuracy than baselines (for the sensitivity, specificity, and precision of these confusion matrices, see \emph{Supplementary materials} Table S1, Table S2, and Table S3). The reason is that, as opposed to Vanilla DNN and LSTM models, FDDLM can extract the signal signatures that contain the long-term memory of the time series. {We also use the convolutional neural network (CNN) model to characterize the dynamics of the physiological signals recorded with sleep monitors to compare it with our FDDLM; the CNN model misclassified $64.49\%$ of the test sets under hold-out validation. For detailed information about the CNN model and testing results, see section \emph{Methods}, subsection \emph{Neural network architecture for the WestRo COPD dataset} and \emph{Supplementary material} section \emph{Training and testing results for CNN})}.

In summary, our model outperforms all baselines in terms of prediction accuracy under both hold-out and $k$-fold cross-validation. The main conclusion is that FDDLM predicts patients' COPD stages with high accuracy and represents an efficient way to detect early COPD stages in suspected individuals. Indeed, such a low-invasive and convenient tool can help physicians make precise diagnoses and provide appropriate treatment plans for suspected patients.

\subsubsection*{Transfer learning}

{To evaluate our models' performance, we utilize the transfer learning mechanism to investigate the generalizability of our FDDLM.  As such, we introduce the WestRo Porti COPD dataset. Transfer learning is a machine learning method that reuses a model designed for analyzing a dataset on another dataset, thus improving the learner from one domain by transferring information from another related domain~\cite{tan2018survey}. The medical subjects in the WestRo Porti are consecutive individuals in the Victor Babes hospital records, screened for sleep apnea with the Porti SleepDoc 7 portable PSG device by recording 6 physiological signals; some individuals are also in various COPD stages. (For detailed information, see \emph{Methods} subsection \emph{Data collection}). The reasons for applying our COPD FDDLM are: (1) we want to verify that our model is valid on an external dataset; (2) we want to test our model's prediction performance when the medical signal records are interfered with by another disease (i.e., sleep apnea). }

{We test FDDLM with the WestRo Porti COPD dataset to check the prediction performance. Since the WestRo COPD Porti dataset only have 6 signals (whereas our model uses 12 signals), we reconstructed the input size of the models from $144\times1$ to $36\times1$, retrained a new FDDLM with WestRo and tested it on the WestRo Porti COPD dataset to check the performance. (Note that the WestRo Porti COPD dataset patients are not included in the WestRo COPD dataset). The prediction accuracy of FDDLM is $90.13\%\pm0.89\%$ with fine-tuning. The explanation for the accuracy drop is that (\textit{i}) the models are previously designed for analyzing medical records with 12 signals, not 6; (\textit{ii}) the two datasets are recorded by two different portable devices having different frequencies, which influences the convergence of the coupling matrices; (\textit{iii}) the co-existed sleep apnea in the medical records gathered from the WestRo Porti COPD dataset also influence the prediction performance. }

\subsubsection*{Summary}

Nowadays, DNN and LSTM models are the two popular deep learning models in analyzing and classifying time-series data. However, they do not present high-performance when the time series are precisely long (the reason is that the current model cannot correctly extract the long-term memory from very long time series). In this work, we developed a novel fractional dynamics-based model which can appropriately analyze long-term memory from COPD physiological signals datasets by extracting fractional features (coupling matrix $A$) from very long time-series data. The extracted fractional features are more straightforward for deep learning models to classify than the raw signals, and even the linear classifier achieves a good accuracy (for detailed information, see "Supplementary materials" subsection "Linear classifier of coupling matrices"). Therefore, based on the results shown in $k$-fold cross-validation, hold-out validation, and external validations, we conclude that our FDDLM has enough generalizability to be applied to different kinds of COPD records that contain physiological signals and is robust enough to eliminate the interference about other diseases. Indeed, based on the transfer learning results, we argue that our FDDLM is robust enough to predict COPD stages across different datasets with high accuracy.

{Besides the high accuracy of our COPD stage prediction method, we made a deeper analysis of the cases where our predictions failed. Overall, we have 19 misclassified cases (out of 534); most of them (i.e., 9) correspond to borderline cases where the spirometry values are exactly (or very close to) threshold values between stages. Some borderline COPD cases also overlap with other respiratory diseases, such as sleep apnea or asthma (3 cases), while one borderline case overlaps with heavy smoking. We also found that 7 misclassified cases overlap with comorbidities, such as severe sleep apnea, asthma, or obesity. For 3 cases, there is no apparent explanation for the misclassification, although one of them is a heavy smoker; in all these 3 cases, the individuals have no COPD but are predicted at stage 2. Another finding is that the noise in physiological signals recordings may rarely cause misclassifications of non-COPD cases (i.e., stage 0), as we noticed in 3 cases (in our datasets, we have 143 patients with stage 0).}

{The overarching conclusion is that comorbidities (especially sleep apnea and asthma) can alter the physiological signals to affect the prediction accuracy, mainly when dealing with borderline COPD stages. Future works have to consider the comorbidity cases carefully. Indeed, more people in the aging population suffer from multi-morbidity, defined as two or more chronic conditions. COPD is common in multi-morbid patients, and many patients with COPD present concomitant other obstructive diseases, such as obstructive sleep apnea (OSA) and asthma, due to an increased prevalence of obesity, smoking, and allergy in the general population~\cite{owens2017overlaps}. Recent estimation of OSA prevalence shows almost 1 billion people affected, with prevalence exceeding 50\% in some countries~\cite{benjafield2019estimation}. Moreover, around 300 million people have asthma worldwide, and it is likely that by 2025 a further 100 million may be affected~\cite{ellwood2020global}.}

\section*{Discussion}

COPD is often a silent and late-diagnosed disease, affecting over 300 million people worldwide; intrinsically, its early discovery and treatment are crucial for the patient's quality of life and---ultimately---survival. 

The inception of a disease entails a preclinical period where it is asymptomatic and---perhaps---reversible; ideally, this period includes very \emph{early} events that can occur even before birth \cite{bolton2015lung}.
Early COPD stages do not exhibit evident clinical signs; therefore, conventional spirometry-based diagnosis becomes improbable. However, the development of biomarkers that include detecting genetic variants for COPD development's susceptibility is a priority. The COPD onset is a phase of early COPD where the disease may express itself with some symptoms, including a minimal airflow limitation \cite{rennard2015early}. In this phase, spirometry is insufficient to attain a reliable diagnosis, which calls for new COPD detection tools. 

The current strategy of waiting for surfacing symptoms to signal the disease presence is not efficient if we want to impact COPD's natural course. Targeting early COPD stages in younger individuals could identify those susceptible to rapid disease progression, leading to novel therapies to alter that progression.
New validated biomarkers (other than spirometry) of different lung function trajectories will be essential for the design of future COPD prevention and treatment trials \cite{rennard2015natural}.
Indeed, spirometry with FEV1 may not be the most sensitive test and may have particular limitations in identifying the early COPD stages. Moreover, impulse oscillometry and specific airway conductance were able to identify more subtle changes in lung function than traditional spirometry \cite{borrill2008use}. Impulse oscillometry can identify abnormalities in patients who report COPD symptoms but do not have abnormal spirometry \cite{frantz2012impulse}.
Such complementary diagnostic modalities could potentially aid in the early recognition of COPD, especially those whose symptoms do not match their spirometry results.

The adjustment of current diagnostic approaches and the adoption of alternative modalities may allow for earlier identification of COPD patients. The period of the most rapid decline in lung function occurs early, and during this period, different testing strategies, smoking cessation efforts, and the initiation of treatment may be most beneficial. 

This work proposes an alternative, precise diagnostic approach to overcome the conventional (spirometry-based) method's limitations by using a fractional dynamics deep learning methodology. We involved the fractional-order dynamics model in extracting the signatures from the physiological signals (recorded by a medical sensor network from suspected patients) and trained a neural network model with these signal signatures to make the COPD stage prediction. From a clinical standpoint, our fluctuation profile analysis for physiological signals with relevance in COPD (see Figure \ref{fig:hyp_test}) shows that multifractality is the fingerprint of healthiness---in healthy people, physiological signals present both short and long-range memory. Conversely, monofractals or narrow-spectrum multifractals indicate a medical condition. We noticed two exceptions to this observation; first, the Pulse signal is narrow-spectrum multifractal even in non-COPD (i.e., stage 0) individuals; second, the Nasal Pressure signal is multifractal in both COPD and non-COPD subjects. The possible explanation for the fact that the Pulse signal is not entirely multifractal in non-COPD individuals is that---most probably---such subjects may have other medical conditions, such as sleep apnea or cardiovascular problems, which (along with the associated medication) altered the short and long-range properties of the signal. (Indeed, all patients were referred for polysomnography because of suspicion of respiratory disorders; some turned out to be COPD-free, yet most of them have other respiratory disorders, such as sleep apnea, as well as cardiovascular conditions.) The Nasal Pressure signal is multifractal in both COPD and non-COPD subjects because it manifests upper airway dynamics, which may be less affected as COPD is an inflammation (resulting in the narrowing) of the lung airways.

We confirm the results with $k$-fold cross-validation and hold-out validation and show that our approach can predict the patients' COPD stages with high accuracy ($98.66\%\pm0.45\%$). The accuracy is particularly high for COPD stages 1--3, suggesting that our method is distinctly efficient for detecting early-stage COPD. Furthermore, based on the transfer learning validation, we prove that our model can also achieve high prediction accuracy when the medical signal records are interfered with by another disease (i.e., sleep apnea). Our work makes two main contributions in medical diagnosis and machine learning fields. First, our fractional dynamics deep learning model makes a precise and robust COPD stage prediction that can work before the disease onset, making it especially relevant for primary care in remote areas or geographical regions missing medical experts, where the vast majority of patients with early and mild COPD are diagnosed and treated. Second, we developed a valid fractional deep learning approach that outperforms the traditional deep learning model (e.g., DNN, LSTM, CNN) of classifying and analyzing very long time-series raw data. (We provide detailed information to explain why our model can efficiently reduce the learning complexity and achieve a high prediction accuracy in section \emph{Methods}, subsection \emph{Mutual information analysis}). 

{Nowadays, the conventional spirometry-based diagnosis is the dominating approach to diagnosing COPD. The problem is that it entails many error-prone steps/stages involving human intervention such that general practitioners or well-trained nurses may also misdiagnose suspected patients (of the 4610 subjects, 96.5\% had a valid screening spirometry test~\cite{soumagne2020quantitative}). Such a result emphasizes that training and technique reinforcement are paramount, yet many primary care units do not have the resources to perform them. In this paper, our fractal dynamics deep learning method eliminates human intervention (and error) as much as possible; any nurse or MD can place the sensors on the patient's body, turn on the NOX device to record the physiological signals in its local memory. Afterward, we are dealing with a completely automated, computer-based process. The sufficient signal length required for a correct diagnostic is 10 minutes (for detailed information, see \emph{Supplementary material}, section \emph{Convergence of coupling matrix}). Therefore, our method is simple, robust, requires little human intervention, and has a relatively small duration of physiological signal records; this also makes it suitable for addressing critical social aspects of healthcare. First, there is equal opportunity in accessing reliable medical consultation for COPD, especially in areas with a lower socioeconomic status where people do not have the means to travel to a specialized state-of-the-art respiratory clinic~\cite{weiss2020global}. With our method, any medical mission in such an area can efficiently record data from many individuals in need and then process it automatically. Second, our method abides by the commandments of universal health care amid the COVID-19 pandemic, as it filters most of the physical interaction entailed by regular spirometry~\cite{chang2016copd}. 
}
\section*{Methods}
\subsection*{Data collection}

\textbf{WestRo COPD dataset:} The study cohort represents consecutive patients from 4 Pulmonology Clinics in Western Romania (i.e., the WestRo cohort, comprising patients from Victor Babe\c{s} -- VB, Medicover 1 -- MD1, Medicover 2 -- MD2, and Cardio Prevent -- CP clinics). Data consist of physiological signals recorded over long periods (i.e., 6-24 hours), using a protocol that ensures complete patient privacy. To obtain a reliable medical diagnostic for each patient, we also collected the following data records: age, sex, body mass index (BMI, as a ratio between mass in kilograms and the squared value of height in meters), smoking history (in years since quitting smoking, with value 0 representing current smokers), FVC and FEV1 in liters and percentage (used to render the COPD stage diagnosis according to the ERS/ATS recommendation \cite{miller2005ats}, with stage 0 representing no COPD), COPD assessment test (CAT) and dyspnea severity with modified Medical Research Scale (MRC) questionnaires, exacerbations (number of moderate to severe exacerbation in the last year), COPD onset (number of years since the onset). For detailed information about CAT and MRC, see \emph{Supplementary material}, section \emph{Standard questionnaires, exacerbation history, and comorbidities of COPD patients}.\\

\noindent
{\textbf{WestRo Porti COPD dataset:} The WestRo Porti cohort consists of polysomnography (PSG) physiological
signals recorded in 13824 medical cases from 534 individuals during 2013-2020. The subjects in the WestRo Porti are consecutive individuals in the Victor Babes hospital records, screened for sleep apnea with the Porti SleepDoc 7 portable PSG device by recording 6 physiological signals (Flow, SpO2, Pulse, Pulsewave, Thorax, Abdomen) overnight, during sleep. The 6 Porti SleepDoc 7 signals correspond, respectively, to the following NOX T3 signals: Flow, Oxygen Saturation Levels, Pulse, Plethysmograph, Thorax Breathing Effort, and Abdomen Breathing Effort.}\\

{In this work, the same medical doctor gave all diagnoses that led to determining the COPD labels across all institutions. Moreover, the medical doctor used the same devices and diagnosis method (sensors to collect physiological signals from patients and spirometers). In addition, the same medical doctor collected the data in all clinics; spirometry was conducted with the help of trained, experienced technicians, certified in pulmonary function testing, following the ATS/ERS protocol (American Thoracic Society/European Respiratory Society). In all clinics included in our study, there is a quality control program for all procedures. }

{Spirometry quality assurance includes examining test values and evaluating both the volume-time and flow-volume curves for evidence of technical errors. During testing, technicians record a valid test composed of at least 3 acceptable maneuvers with consistent (i.e., repeatable) results for FVC and FEV1. Achieving repeatability during testing means that the differences between the largest and second-largest values for both FVC and FEV1 are within 150 ml. Additional maneuvers can be attempted---up to a maximum of 8---to meet these criteria for a valid test. The observer bias is reduced by ensuring that observers are well trained (specialized clinics do that regularly with certification diplomas), having clear rules and procedures in place for the experiment (i.e., the ERS/ATS protocol), and ensuring that behaviors are clearly defined. Therefore, since the same medical doctor performed all evaluations with the same equipment and diagnosis approach, we are confident that we substantially mitigated the intra- and inter-observer variability.}

\subsection*{Multifractal detrended fluctuation analysis}

Multifractal detrended fluctuation analysis (MF-DFA) is an effective approach to estimate the multifractal properties of biomedical signals\cite{ihlen2012introduction}. The first step of MF-DFA is to calculate the cumulative profile ($Y(t)$)
\begin{equation}
Y(t)=\sum_{i=1}^{t} X(i)-\langle X(i)\rangle ,
\label{eqn:cumulative sum}
\end{equation}
where $X$ is a bounded time series. Then, divide the cumulated signal equally into $N_{s}$ non-overlapping time windows with length $s$, and remove the local line trend (local least-squares straight-line fit) $y_{v}$ from each time window. Therefore, $F(v,s)$ characterizes the root-mean-square deviation from the trend (i.e., the fluctuation), 
\begin{equation}
F(v,s)=\sqrt{\frac{1}{s}\sum_{i=1}^{s}{\{Y\lfloor(v-1)s+i\rfloor-y_{v}(i)\}^{2}}}.
\label{eqn:f(v,s)}
\end{equation}
In \cite{mukli2015multifractal}, the authors defined the scaling function as
\begin{equation}
S(q,s)=\bigg\{\frac{1}{N_{s}}\sum_{v=1}^{N_{s}}{\mu(v,s)}^{q} \bigg\}^{1/q},
\label{eqn:s(s)}
\end{equation}
where $\mu$ is an appropriate measure which depends on the scale of the observation ($s$). Hence, the scaling function is defined by substituting equation \ref{eqn:f(v,s)} into equation \ref{eqn:s(s)},
\begin{equation}
S_{F}(q,s)=\bigg\{\frac{1}{N_{s}}\sum_{v=1}^{N_{s}}{ \Big\{\frac{1}{s}\sum_{i=1}^{s}{\{Y\lfloor(v-1)s+i\rfloor-y_{v}(i)\}^{2}}\Big\}^{q/2}}  \bigg\}^{1/q}.
\label{eqn:s(qs)}
\end{equation}
The moment-wise scaling functions for a multifractal signal exhibit a convergent structure that yields to a focus point for all $q$-values. Such focus points\cite{mukli2015multifractal} can be deduced from equation \ref{eqn:s(s)}, by considering the signal length as a scale parameter,
\begin{equation}
S(q,L)=\bigg\{\frac{1}{N_{s}}\sum_{v=1}^{N_{L}}{\mu(v,L)}^{q} \bigg\}^{1/q}=\{\mu(v,L)^{q}\}^{1/q}=\mu(v,L),
\label{eqn:s(l)}
\end{equation}
where the value of $\mu$ represents the entire signal, namely $N_{L}=1$ (i.e., takes only one time window into consideration). According to equation \ref{eqn:s(l)}, the scaling function $S(q,L)$ becomes independent from the exponent $q$ and the moment-wise scaling functions will converge to $\mu(v,L)$ which is the mathematical definition of the force point.

\subsection*{Neural network architecture for the WestRo COPD dataset}
{\textbf{Fractional dynamics deep learning model (FDDLM).} 
In our work, FDDLM consists of two parts: (1) fractional signature extraction (for more details, please see section \emph{Methods}, subsection \emph{Multifractal detrended fluctuation analysis}) and (2) a deep learning model. Keeping in mind the input size of our training data (i.e., the coupling matrix $A$) and available GPU computational power, we constructed a deep neural network (DNN) architecture to handle the training and prediction progress. We built the network with the TensorFlow Python framework~\cite{abadi2016tensorflow}. Our deep neural network consists of 6 layers: 1 input layer, 2 hidden layers, 2 dropout layers, and 1 output layer. Also, we resampled the input data (matrix $A$) to $144\times1$ voxels and normalized each value within the range [0, 1] (normalization is a technique for training deep neural networks that standardizes the inputs to a layer). We placed the dropout layers after each hidden layer with a 20\% drop rate (the first hidden layer has 300 neurons and the second hidden layer has 100 neurons); each fully connected hidden layer utilizes the $ReLU$ activation function. The $softmax$ is utilized as the activation function in the output layer. The DNN is optimized with the $rmsprop$ optimizer with a learning rate of 0.0001 and trained with the $crossentropy$ loss function. FDDLM is trained over 500 epochs with a batch size of 64 samples. Overall, the number of trainable parameters of the deep learning model is 74,105.}\\

\noindent
{\textbf{Vanilla deep neural network (DNN) model.} 
The Vanilla DNN model shares the same network structure with the deep learning model in our FDDLM, except the input layer. The Vanilla DNN contains 6 layers: 1 input layer (the input data, namely, the physiological signals are reshaped to $72000\times1$ voxels, and each value is normalized within the range [0, 1]), 2 hidden layers (the first hidden layer has 300 neurons and the second hidden layer has 100 neurons), 2 dropout layers, and 1 output layer. The activation function for each fully connected hidden layer is $ReLU$, and the activation function for the output layer is $softmax$. The Vanilla DNN model is optimized with the $rmsprop$ optimizer, having a default learning rate of 0.0001, and trained with the $crossentropy$ loss function. The model is trained over 500 epochs with a batch size of 64 samples. The total number of trainable parameters of the Vanilla DNN model is 21,630,905.}\\

\noindent
{\textbf{Long short-term memory (LSTM) model.} The LSTM model in this work has the following layers: an input layer (the input physiological signals are reshaped to $6000\times12$ voxels, and each value is normalized within the interval [0, 1]), an LSTM layer (with 300 neurons), a dropout layer (with a 0.2 dropout rate), a dense layer (with 100 neurons), a dropout layer (with a 0.2 dropout rate), and an output layer. $ReLU$ is the activation function for the LSTM and dense layers. The model is optimized with $rmsprop$ having a default learning rate of 0.0001 and trained with the $crossentropy$ loss function. The LSTM model is trained over 500 epochs with a batch size of 64 samples. The total number of trainable parameters of the LSTM model is 535,805.}\\

\noindent
{\textbf{Convolutional neural network (CNN) model.} The CNN model in this paper has the following layers: an input layer (the input physiological signals are reshaped to $72000\times1$ voxels, each value normalized within the range [0, 1]), a convolutional layer (64 neurons), a flatten layer, a dropout layer (with a 0.2 dropout rate), a dense layer (with 32 neurons), a dropout layer (with a 0.2 dropout rate), and an output layer (with 5 neurons). $ReLU$ is the activation function for the convolutional and dense layers, while $softmax$ is the activation function for the output layer. The CNN model is optimized with $rmsprop$ having a default learning rate of 0.0001 and trained with the $crossentropy$ loss function. The CNN model is trained over 500 epochs with a batch size of 64 samples; the total number of trainable parameters is 147,456,453.}\\

\begin{figure}[!ht]
\begin{floatrow}
  \begin{adjustbox}{width=0.45\textwidth}
    \ffigbox{%
      {\includegraphics[width=
    0.45\textwidth]{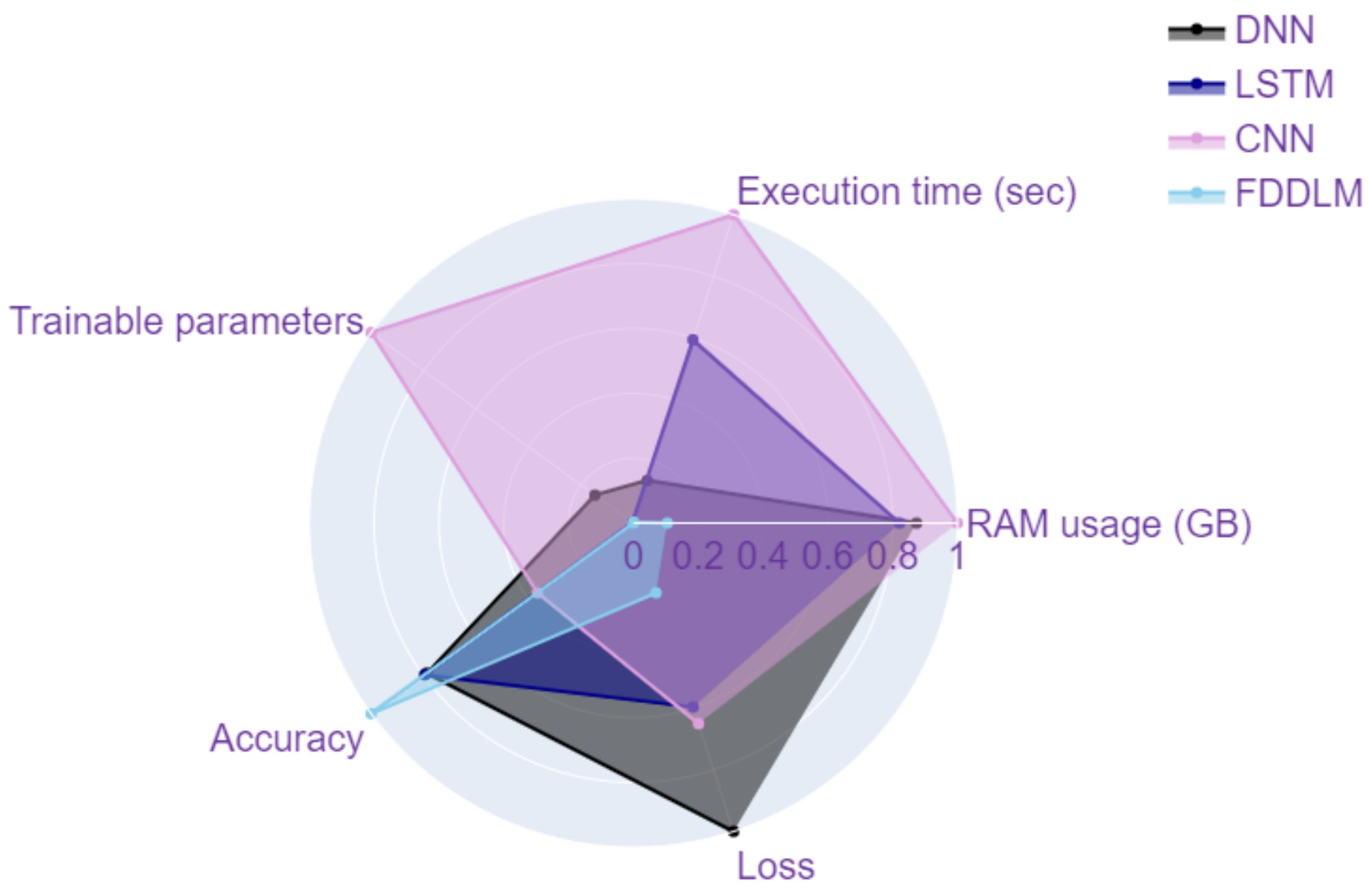}}
    }
    {%
    \caption{{{\textbf{The radar plot for measuring the complexity and prediction performance for the WestRo COPD dataset across different deep learning models under $k$-fold validation ($k=5$):} fractional-dynamics deep learning model (FDDLM), Vanilla deep neural network (DNN), long short-term memory (LSTM), and convolutional neural network (CNN). We normalized all the values represented in this plot.
}}}%
    \label{fig:radar}
    }
\end{adjustbox}

\capbtabbox{%
  \begin{adjustbox}{width=0.53\textwidth}
  \begin{tabular}{ccccc}
\toprule
  & FDDLM & DNN & LSTM & CNN\\ \hline
RAM usage (GB) $\downarrow$    & \textbf{0.98} & 8.17 & 7.68 & 9.35 \\
Execution time (sec) $\downarrow$   & \textbf{1076} & 47,590 & 205,135& 345,085 \\ 
Trainable parameters $\downarrow$    & \textbf{74,105} & 21,630,905 & 535,805 & 147,456,453\\ 
Test accuracy (\%) $\uparrow$   & \textbf{98.66} & 77.72 & 78.54 & 36.12\\
Loss $\downarrow$   & \textbf{0.2170} & 0.9601 & 0.5728 & 0.6245\\ 
\bottomrule
\end{tabular}
\end{adjustbox}
}{%
  \caption{{{\textbf{Complexity and prediction performance for the WestRo COPD dataset across different deep learning models under $k$-fold validation ($k=5$):} fractional dynamics deep learning model (FDDLM), Vanilla deep neural network (DNN), long short-term memory (LSTM), and convolutional neural network (CNN). $\uparrow/\downarrow$ indicates higher/lower values are better. All results are evaluated on the same machine for fair comparison. 
}}}
  \label{tb:radar}
}
\end{floatrow}
\end{figure}

{We further compare resource usage and performance across different models under $k$-fold cross-validation---namely, FDDLM, Vanilla DNN, LSTM, and CNN---by measuring the following metrics: execution time, trainable parameters, RAM usage (in GB) and accuracy. The evaluation results are shown in Figure~\ref{fig:radar} and Table~\ref{tb:radar} (Of note, the values in Table~\ref{tb:radar} are the mean results under $k$-fold validation ($k=5$) and the values in Figure~\ref{fig:radar} are the normalized results from Table~\ref{tb:radar}). For Figure~\ref{fig:radar} and Table~\ref{tb:radar}, we observe that our FDDLM has the highest accuracy (i.e., 96.18\%) with the lowest complexity (i.e., memory usage and execution time). These findings indicate that our model's predictions are more accurate while requiring a lower complexity than traditional machine learning models (i.e., Vanilla DNN, CNN, and LSTM).
}

\subsection*{{Challenges and limitations of spirometry in COPD}}
{Spirometry is a physiological test that measures the maximal air volume that an individual can inspire and expire with maximal effort, thus assessing the effect of a disease on lung function. Together with the medical history, symptoms, and other physical findings, it is an essential tool that provides essential information to clinicians in reaching a proper diagnosis~\cite{graham2019standardization}. Indeed, standard spirometry is a laborious procedure: it needs preparation, a bronchodilation test, performance assurance, and evaluation.~\cite{pauwels2001global}}\\

\noindent
{\textbf{Preparation: }(1) The ambient temperature, barometric pressure, and time of day must be recorded. (2) Spirometers are required to meet International Organization for Standardization (ISO) 26782 standards, with a maximum acceptable accuracy error of $\pm2.5\%$. (3) Spirometers need calibration daily, with calibration verification at low, medium, and high flow. (4) The technicians have to make sure that the device produces a hard copy of the expiratory curve plot to detect common technical errors. (5) The pulmonary function technician needs training in the optimal technique, quality performance, and maintenance. (6) There are activities that patients should avoid before testing, such as smoking or physical exercise. (7) Patients should be adequately instructed and then supported to provide a maximal effort in performing the test to avoid underestimating values and ultimately diagnosis errors.}\\

\noindent
{\textbf{Bronchodilation: }(1) The forced expiratory volume in one second (FEV1) should be measured 10-15 minutes after the inhalation of 400 mcg short-acting beta2 agonist, or 30-45 minutes after 160 mcg short-acting anticholinergic, or the two combined~\cite{miller2005standardization}. (2) Physicians also developed new withholding times for bronchodilators before bronchodilator responsiveness testing~\cite{graham2019standardization}.}\\

\noindent
{\textbf{Performance assurance: }(1) Spirometry should be performed using standard techniques. (2) The expiratory volume/time traces should be smooth and without irregularities, with $a$ less than 1 second pause between inspiration and expiration. (3) The recording should be long enough to reach a volume plateau; it may take more than 15 seconds in severe cases~\cite{glover2014forced}. (4) Both forced vital capacity (FVC) and FEV1 should represent the biggest value obtained from any of three out of a maximum of eight technically good curves, and the values should vary by no more than 5\% or 150 ml---whichever is bigger~\cite{enright2004repeatability}. (5) The FEV1/FVC ratio should be taken as the technically acceptable curve with the largest sum of FVC and FEV1~\cite{hankinson2015use}.}\\

\noindent
{\textbf{Evaluation: }(1) The measurements evaluation compares the results with appropriate reference values---specific to each age, height, sex, and race group. (2) The presence of a post-bronchodilator FEV1/FVC < 0.70 confirms the presence of airflow limitation~\cite{barjaktarevic2018bronchodilator}.}\\

\noindent
{It is clear that the diagnosis process---primarily relying on spirometry---is pretty complex and, thus, prone to errors because of human intervention. The large university clinics, such as our Victor Babes clinic in Timisoara (and the other institutions included in our paper's recordings), avoid errors by carefully training their personnel and enforcing strict procedures. Additionally, experienced and well-trained physicians corroborate the spirometry results with other clinical data, such that diagnostic mistakes are highly improbable. However, we face a big problem with spirometry in primary care offices, which do not have all resources to consistently abide by the quality assurance steps (preparation, bronchodilation test, performance assurance, and evaluation). Hegewald ML et al. showed that most spirometers tested in primary care offices were not accurate, and the magnitude of the errors resulted in significant changes in the categorization of patients with COPD. Indeed, they obtained acceptable quality tests for only 60\% of patients~\cite{hegewald2016accuracy}. In a similar study, the authors reported a spirometry accuracy varying from  69.1\% to 81.4\% in the primary care offices~\cite{hegewald2015accuracy}. These prior experimental studies and findings are significant for the medical community and constitute the motivation for our paper since primary care offices have an essential role in the early detection of COPD cases. }

\subsection*{{Definition of COPD stages}}
{The diagnosis of COPD is based on persistent respiratory symptoms such as cough, sputum production, and dyspnea, together with airflow limitation (caused by significant exposure to smoking, noxious particles, or gases) evaluated with spirometry. The labels or disease stages are defined in the standard guideline of the worldwide medical community \cite{rabe2007global}. Based on the FEV1 (forced expiratory volume in one second) value measured by spirometry, the Global Initiative for Chronic Obstructive Lung Disease (GOLD) guideline system categorizes airflow limitation into stages. In patients with FEV1/FVC (forced vital capacity) <0.70, the standard labels: (1) STAGE 1 -- mild: FEV1$\geq80\%$; (2) STAGE 2 -- moderate: $50\%\leq$ FEV1 <80\%; (3) STAGE 3 -- severe: $30\% \leq$ FEV1 $<50\%$; (4) STAGE 4 -- very severe: FEV1 <30\%. Additionally, in this paper, we assign the STAGE 0 label to patients without COPD (i.e., FEV1/FVC$\geq$0.70).
}

\subsection*{{Early COPD stages}}
{Nowadays, there have been many debates in the literature regarding the early stage of COPD or the so-called asymptomatic COPD. Patients with COPD often underestimate the severity of the disease—--primarily early morning and nighttime symptoms. The reasons may be the slow onset of their symptoms, cough due to a long cigarette smoking history, and dyspnea attributed to getting older. The majority of patients from a European cohort stated that they were not wholly frank with their doctors during visits when reporting their symptoms and quality of life ~\cite{celli2017perception}. }

{Around 36\% of patients who describe their symptoms as mild-to-moderate also admit to being too breathless to leave the house. For these reasons, there are two validated questionnaires (i.e., CAT and Modified Medical Research Council (mMRC)) that allow clinicians to accurately and objectively assess COPD symptoms. CAT is a globally used, 8 question, patient-filled questionnaire to evaluate the impact of COPD (cough, sputum, dyspnea, chest tightness) on health status. The range of CAT scores is 0–40. Higher scores denote a more severe impact of COPD on a patient's life~\cite{jones2016copd}. The mMRC Dyspnea Scale stratifies dyspnea severity in respiratory diseases, particularly COPD; it provides a baseline assessment of functional impairment attributable to dyspnea in respiratory diseases. Moreover, despite being highly symptomatic (mMRC$\geq$2 and CAT$\geq$10) and having at least one exacerbation, many COPD patients did not seek medical help, as they felt COPD symptoms as part of their daily smoking routine or due to aging. COPD awareness is poor among smokers; the smoker population underestimates their respiratory symptoms, while their exercise activity is reduced many times. Not surprisingly, 14.5\% of the newly diagnosed COPD population was reported as asymptomatic in primary care clinics~\cite{gogou2021underestimation}. Also, there is a high prevalence of COPD among smokers with no symptoms~\cite{sansores2015prevalence}. We did not consider subjectively reported or observed clinical symptoms; instead, our analysis is based only on objectively measured parameters (i.e., physiological signals).}

{Spirometry as a screening tool for the early stage of the disease is not entirely robust~\cite{bibbins2016statin}. Indeed, spirometry can diagnose asymptomatic COPD, but its use is only recommended in smokers or individuals with a history of exposure to other noxious stimuli~\cite{soriano2009screening}. Despite having an apparent normal lung function, smokers with normal spirometry but a low diffusing capacity of the lung for carbon monoxide (DLCO) are at significant risk of developing COPD with obstruction to airflow ~\cite{sanchez2014disease}---a category that may also be asymptomatic COPD. Moreover, no other disease markers are known to date to predict which patients with COPD of recent onset will progress to more significant disease severity. }

{Nonetheless, undiagnosed asymptomatic COPD has an increased risk of exacerbations and pneumonia. For these reasons, we need better initiatives for the early diagnosis and treatment of COPD~\cite{ccolak2017prognosis}. Our method also aims at addressing the problem of early detection because it has an excellent accuracy at detecting early stages 1 and 2, which can also be detected with Spirometry. However, if our method can identify asymptomatic COPD that spirometry-based methods cannot see remains an open question; to that end, we need a longitudinal study starting with a significant cohort, which tracks the evolution of individuals over time to see if those predicted as asymptomatic COPD indeed develop the symptomatic form of the disease after several years. }

\section*{Acknowledgements}

P.B., C.Y., and G.G. gratefully acknowledge the support by the National Science Foundation under the Career Award CPS/CNS-1453860, the NSF award under Grant numbers CCF-1837131, MCB-1936775, and CNS-1932620, the U.S. Army Research Office (ARO) under Grant No. W911NF-17-1-0076 and the DARPA Young Faculty Award and DARPA Director Award, under grant number N66001-17-1-4044, and a Northrop Grumman grant. M.U. A.L., L.U. and S.M. gratefully acknowledge that this work was supported by a grant of the Romanian National Authority for Scientific Research and Innovation, CNCS/CCCDI--UEFISCDI, project number PN-III-P2-2.1-PED-2016-1145, within PNCDI III project number 31 PED/2017: INCEPTION. There was no additional external funding received for this study. The funder had no role in study design, data collection and analysis, decision to publish, or preparation of the manuscript. The views, opinions, and/or findings contained in this article are those of the authors and should not be interpreted as representing official views or policies, either expressed or implied by the Defense Advanced Research Projects Agency, the Department of Defense or the National Science Foundation.

\section*{Author contribution statement}

M.U., L.U., P.B., and S.M. contributed to proposing the idea. C.Y., G.G., and A.L. contributed to analyze the fractional-order dynamic characteristics of physiological signals. C.Y. and A.L. designed the DNN model and performed the experiments. M.U., L.U., and S.M. collected experimental data. C.Y., P.B., and M.U. contributed to write the manuscript. All authors provided feedback on the manuscript. 

\section*{Additional information}
\subsection*{Code and data availability}
The code for reproducing the results is provided on:
{\href{https://github.com/chenzhoy/Fractional-dynamics-foster-deep-learning-of-COPDstage-prediction.git}{https://github.com/chenzhoy/Fractional-dynamics-foster-deep-learning-of-COPDstage-prediction.git}}
\subsection*{Competing Interests}
The Authors declare no Competing Financial or Non-Financial Interests.

\bibliography{sample}
\end{document}


\flushbottom
\maketitle
\section*{Supplementary information}
\subsection*{{Sensitivity, specificity, and precision rate of the confusion matrices}}
{The WestRo COPD dataset consists of physiological signals recorded over consecutive patients from four Pulmonology Clinics in
Western Romania (Victor Babe\c{s} Hospital – VB, Medicover 1 – MD1, Medicover 2 – MD2, and Cardio Prevent – CP clinics). This supplementary material displays detailed results about the confusion matrices presented in the manuscript (Figure 7). We generated the confusion matrices with our fractional dynamics deep learning model (FDDLM), the vanilla DNN model, and the LSTM model and by holding out data gathered at each institution---at a time---as test sets.  (The vanilla DNN and LSTM models have a similar network structure with the fractional dynamics deep learning model, except the input size.)  We analyze the sensitivity, specificity, and precision of these confusion matrices to measure the performance of each machine learning model. We present the results in Tables \ref{tab:12}, \ref{tab:2}, and \ref{tab:3}. The results in these tables show that, despite LSTM and vanilla DNN models having high specificity, the prediction sensitivity and precision of these two models are not good (with the worst sensitivity
values dropping to 36.77\% and 35.33\%). In contrast, our fractional dynamics deep learning model shows relatively high values across the three performance measurements. We conclude that our model predicts patients' COPD stage with higher accuracy than vanilla DNN and LSTM models---without overfitting---therefore representing an effective alternative to the spirometry-based diagnostic.}

\begin{table}[]
\centering
\caption{The COPD stage prediction results with the fractional dynamics deep learning model by holding out data gathered from each institution at a time as the test set.  }
\resizebox{16cm}{!}{%
\begin{tabular}{ccccccc}
\toprule
Medical institution & Parameters & Stage 0 & Stage 1 & Stage 2 & Stage 3 & Stage 4 \\ \hline
\multirow{3}{*}{VB} & Sensitivity & 97.96\% & 86.45\% & 99.99\% & 99.62\% & 99.99\% \\ 
 & specificity & 99.99\% & 99.99\% & 97.01\% & 99.88\% & 99.79\% \\ 
 & Precision & 99.99\% & 99.99\% & 94.42\% & 99.62\% & 98.84\% \\ \hline
\multirow{3}{*}{MD1} & Sensitivity & 99.99\% & 98.76\% & 94.77\% & 97.12\% & 95.86\% \\ 
 & specificity & 98.85\% & 98.10\% & 99.71\% & 99.33\% & 99.99\% \\ 
 & Precision & 93.29\% & 89.89\% & 99.52\% & 97.12\% & 99.99\% \\ \hline
\multirow{3}{*}{MD2} & Sensitivity & 98.73\% & 98.20\% & 86.31\% & 94.51\% & 99.34\% \\ 
 & specificity & 96.70\% & 96.00\% & 98.70\% & 99.77\% & 99.78\% \\ 
 & Precision & 83.87\% & 82.00\% & 99.44\% & 98.85\% & 98.68\% \\ \hline
\multirow{3}{*}{CP} & Sensitivity & 95.60\% & 95.83\% & 99.14\% & 99.35\% & 99.52\% \\ 
 & specificity & 99.99\% & 99.66\% & 99.10\% & 99.23\% & 99.74\% \\ 
 & Precision & 99.99\% & 97.87\% & 98.29\% & 97.47\% & 98.56\% \\ 
   \bottomrule
\end{tabular}%
}

\label{tab:12}
\end{table}

\begin{table}[]
\centering
\caption{The COPD stage prediction results with the vanilla DNN model by holding out data gathered from each institution at a time as the test set. 
}
\resizebox{16cm}{!}{%
\begin{tabular}{ccccccc}
\toprule

Medical institution & Parameters & Stage 0 & Stage 1 & Stage 2 & Stage 3 & Stage 4 \\ \hline
\multirow{3}{*}{VB} & Sensitivity & 91.16\% & 36.77\% & 94.09\% & 89.02\% & 74.71\% \\ 
 & specificity & 94.90\% & 96.33\% & 93.34\% & 94.91\% & 96.91\% \\ 
 & Precision & 73.22\% & 61.96\% & 87.72\% & 84.53\% & 81.41\% \\ \hline
\multirow{3}{*}{MD1} & Sensitivity & 79.61\% & 54.04\% & 86.62\% & 72.60\% & 77.24\% \\ 
 & specificity & 96.75\% & 90.48\% & 93.54\% & 96.32\% & 94.07\% \\ 
 & Precision & 79.60\% & 49.15\% & 89.88\% & 82.07\% & 66.27\% \\ \hline
\multirow{3}{*}{MD2} & Sensitivity & 97.25\% & 51.04\% & 92.67\% & 94.52\% & 81.64\% \\ 
 & specificity & 96.59\% & 96.64\% & 93.93\% & 96.36\% & 98.52\% \\ 
 & Precision & 81.57\% & 71.53\% & 88.84\% & 88.51\% & 68.93\% \\ \hline
\multirow{3}{*}{CP} & Sensitivity & 78.48\% & 52.09\% & 88.51\% & 97.80\% & 79.61\% \\ 
 & specificity & 96.81\% & 91.01\% & 92.11\% & 96.61\% & 99.56\% \\ 
 & Precision & 81.04\% & 51.78\% & 87.44\% & 85.57\% & 96.80\% \\ 
  \bottomrule
\end{tabular}%
}

\label{tab:2}
\end{table}

\begin{table}[]
\centering
\caption{The COPD stage prediction results with the LSTM model by holding out data gathered from each institution at a time as the test set. }
\resizebox{16cm}{!}{%
\begin{tabular}{ccccccc}
\toprule

Medical institution & Parameters & Stage 0 & Stage 1 & Stage 2 & Stage 3 & Stage 4 \\ \hline
\multirow{3}{*}{VB} & Sensitivity & 99.99\% & 50.63\% & 80.91\% & 87.12\% & 88.82\% \\ 
 & specificity & 91.90\% & 92.55\% & 96.89\% & 99.99\% & 96.81\% \\ 
 & Precision & 65.33\% & 52.98\% & 92.90\% & 99.99\% & 83.43\% \\ \hline
\multirow{3}{*}{MD1} & Sensitivity & 75.16\% & 73.29\% & 78.91\% & 73.07\% & 82.76\% \\ 
 & specificity & 96.23\% & 86.16\% & 98.95\% & 97.22\% & 94.18\% \\ 
 & Precision & 76.16\% & 47.39\% & 98.02\% & 85.87\% & 68.18\% \\ \hline
\multirow{3}{*}{MD2} & Sensitivity & 66.46\% & 35.33\% & 97.07\% & 70.72\% & 93.42\% \\ 
 & specificity & 98.42\% & 93.19\% & 86.26\% & 98.98\% & 93.31\% \\ 
 & Precision & 87.50\% & 47.97\% & 80.86\% & 94.57\% & 68.93\% \\ \hline
\multirow{3}{*}{CP} & Sensitivity & 99.99\% & 53.13\% & 77.80\% & 89.35\% & 99.03\% \\ 
 & specificity & 92.32\% & 91.14\% & 99.99\% & 99.80\% & 97.13\% \\ 
 & Precision & 66.91\% & 49.76\% & 99.99\% & 99.28\% & 86.13\% \\ 
 \bottomrule
\end{tabular}%
}

\label{tab:3}
\end{table}
\begin{figure}[h!]
\centering
\includegraphics[width=1\textwidth]{sfigs/cnn_results.pdf}
\caption{{\textbf{Training and testing results in terms of accuracy, loss, and AUC of the convolutional neural network model for the $k$-fold cross-validation and hold out validation.} Training/testing accuracy (a), loss (b), and AUC (c) for convolutional neural network model under $k$-fold cross-validation, where the training processes utilize physiological signals. Training/testing accuracy (d), loss (e), and AUC (f) for convolutional neural network model under hold-out validation, where the training processes utilize physiological signals. }}
\label{fig:cnn}
\end{figure}
\subsection*{{Convolutional neural network (CNN) training and testing results}}
{In figure~\ref{fig:cnn}, panels (a), (b), and (c) illustrate the training and testing results in terms of accuracy, loss, and AUC for the CNN model trained with physiological signals (raw data) under $k$-fold cross-validation ($k=5$). The testing accuracy of CNN indicates a lower predicting accuracy (36.12\%$\pm$0.001\%) compared with our fractional dynamics model (98.66\%$\pm$0.447\%). In figure~\ref{fig:cnn}, panels (d), (e), and (f) present the training and testing results in terms of accuracy, loss, and AUC for the CNN model trained with physiological signals (raw data) under hold-out validation. The testing accuracy under hold-out validations is 35.51\%$\pm$0.009\%, significantly lower than our model’s predicting accuracy (95.88\%$\pm$1.76\%). Hence, we conclude that our fractional-dynamics deep learning model predicts patients’ COPD stages with higher accuracy than the vanilla DNN, LSTM, and CNN models trained with physiological signals.} 

\subsection*{{Hurst exponents of physiological signals}}
{In this section, we show the Hurst exponents in non-derived signals (Thorax, Oxygen Saturation, Pulse, Plethysmograph, Nasal Pressure, and Abdomen) among COPD patients for the intermediate-stage patients ($q\in[-5, 5]$). The Hurst exponent measures the long-term memory of time series; different Hurst exponent values reveal different evolving variations in time series with different fractal features. Figure~\ref{fig:hq1} presents the $H(q)$ confidence intervals for Abdomen, Thorax, and Pulse signals, for different $q$ values, across different COPD stages (i.e., 0, 1, 2, 3) with 95\% confidence intervals; Figure~\ref{fig:hq2} presents the $H(q)$ confidence intervals for Nasal Pressure, Oxygen Saturation, and Plethysmograph signals, for different $q$ values, across different stages (0, 1, 2, 3) with 95\% confidence intervals. In both Figure~\ref{fig:hq1} and Figure~\ref{fig:hq2}, the $H(q)$ of physiological signals extracted from healthy people (stage 0) are plotted as purple patterns in each panel for reference. From these two figures, we notice that different stages have different $H(q)$ confidence intervals under the same $q$ values, which is the  evidence that all physiological signals extracted from patients with different COPD stages have different fractional dynamic characteristics (for reference, Figure~\ref{fig:all1} and Figure~\ref{fig:all2} present the comparison of terms of $H(q)$ intervals for COPD patients for different $q$ values ($q\in \left\{-5, 0, 5\right\}$). Hence, it makes sense to analyze the spatial coupling between these physiological processes (signals) across time to investigate the different fractional features across signals recorded from different COPD patients. }\\
Figure~\ref{fig:hyp_all} to show the scaling functions of physiological signals extracted from all stage 4 patients and the healthy people (stage 0) in our dataset (where $q\in[-5, 5]$); the purple areas are the confidence intervals.
\response{We also used the MF-DFA toolbox provided by~\cite{mukli2015multifractal} to recompute the multifractal scaling functions of the signals collected from stage 0 and 4 participants (we choose the same participants in Figure 2 for reanalyzing the $S(q)$). The results are shown in Figure~\ref{fig:mu}}

\begin{figure}
\centering
\includegraphics[width=1\textwidth]{sfigs/hq_curves1.pdf}
\caption{{\textbf{Multifractal analysis of 3 non-derived physiological signals (Abdomen, Thorax, and Pulse) extracted from non-COPD individuals (i.e., stage 0) and stage 4 COPD patients with 95\% confidence interval.}We present the generalized Hurst exponent ($H(q)$) as a function of $q$-th order moments ($q\in[-5, 5]$) for Abdomen (a), Thorax (d), and Pulse (g) signals extracted from stage 1 patients; $H(q)$ as a function of $q$-th order moments ($q\in[-5, 5]$) for Abdomen (b), Thorax (e), and Pulse (h) signals extracted from stage 2 patients; $H(q)$ as a function of $q$-th order moments  ($q\in[-5, 5]$) for Abdomen (c), Thorax (f), and Pulse (i) signals extracted from stage 3 patients. For reference, the $H(q)$ function for all physiological signals (Abdomen, Thorax, and Pulse) extracted from the healthy people (stage 0) are plotted in (a-i) panels as purple curves.}}
\label{fig:hq1}
\end{figure}

\begin{figure}
\centering
\includegraphics[width=1\textwidth]{sfigs/hq_curves2.pdf}
\caption{{\textbf{Multifractal analysis of 3 non-derived physiological signals (Nasal Pressure, Oxygen Saturation, and Plethysmograph) extracted from non-COPD individuals (i.e., stage 0) and stage 4 COPD patients with 95\% confidence interval.} We present the generalized Hurst exponent ($H(q)$) as a function of $q$-th order moments ($q\in[-5, 5]$) for Nasal Pressure (a), Oxygen Saturation (d), and Plethysmograph (g) signals extracted from stage 1 patients; $H(q)$ as a function of $q$-th order moments ($q\in[-5, 5]$) for Nasal Pressure (b), Oxygen Saturation (e), and Plethysmograph (h) signals extracted from stage 2 patients; $H(q)$ as a function of $q$-th order moments  ($q\in[-5, 5]$) for Nasal Pressure (c), Oxygen Saturation (f), and Plethysmograph (i) signals extracted from stage 3 patients. For reference, the $H(q)$ function for all physiological signals (Nasal Pressure, Oxygen Saturation, and Plethysmograph) extracted from the healthy people (stage 0) are plotted in the (a-i) panels as purple curves.}}
\label{fig:hq2}
\end{figure}

\begin{figure}
\centering
\includegraphics[width=1\textwidth]{sfigs/all_stages_1.pdf}
\caption{{\textbf{Comparison in multifractal features between 3 non-derived physiological signals extracted from patients with different COPD stages.} We plot the generalized Hurst exponent ($H(q)$) for Abdomen signals extracted from COPD patients in different stages under different $q$-th order moments ($q = -5$ (a), $q = 0$ (b), and $q = 5$ (c)); Generalized Hurst exponent ($H(q)$) for Thorax signals extracted from COPD patients with different stages under different $q$-th order moments ($q = -5$ (d), $q = 0$ (e), and $q = 5$ (f));  $H(q)$ for Pulse signals extracted from COPD patients in different stages under different $q$-th order moments ($q = -5$ (g), $q = 0$ (h), and $q = 5$ (i)). }}
\label{fig:all1}
\end{figure}

\begin{figure}
\centering
\includegraphics[width=1\textwidth]{sfigs/all_stages2.png}
\caption{{\textbf{Comparison of multifractal features between 3 non-derived physiological signals extracted from patients with different COPD stages.} We plot the generalized Hurst exponent ($H(q)$) for Nasal Pressure signals extracted from COPD patients in different stages under different $q$-th order moments ($q=-5$ (a), $q=0$ (b), and $q=5$ (c)); $H(q)$ for Oxygen Saturation signals extracted from COPD patients with different stages under different $q$-th order moments ($q=-5$ (d), $q=0$ (e), and $q=5$ (f));  $H(q)$ for Plethysmograph signals extracted from COPD patients in different stages under different $q$-th order moments ($q=-5$ (g), $q=0$ (h), and $q=5$ (i)).}}
\label{fig:all2}
\end{figure}

\begin{figure}[h!]
\centering
\includegraphics[width=\textwidth]{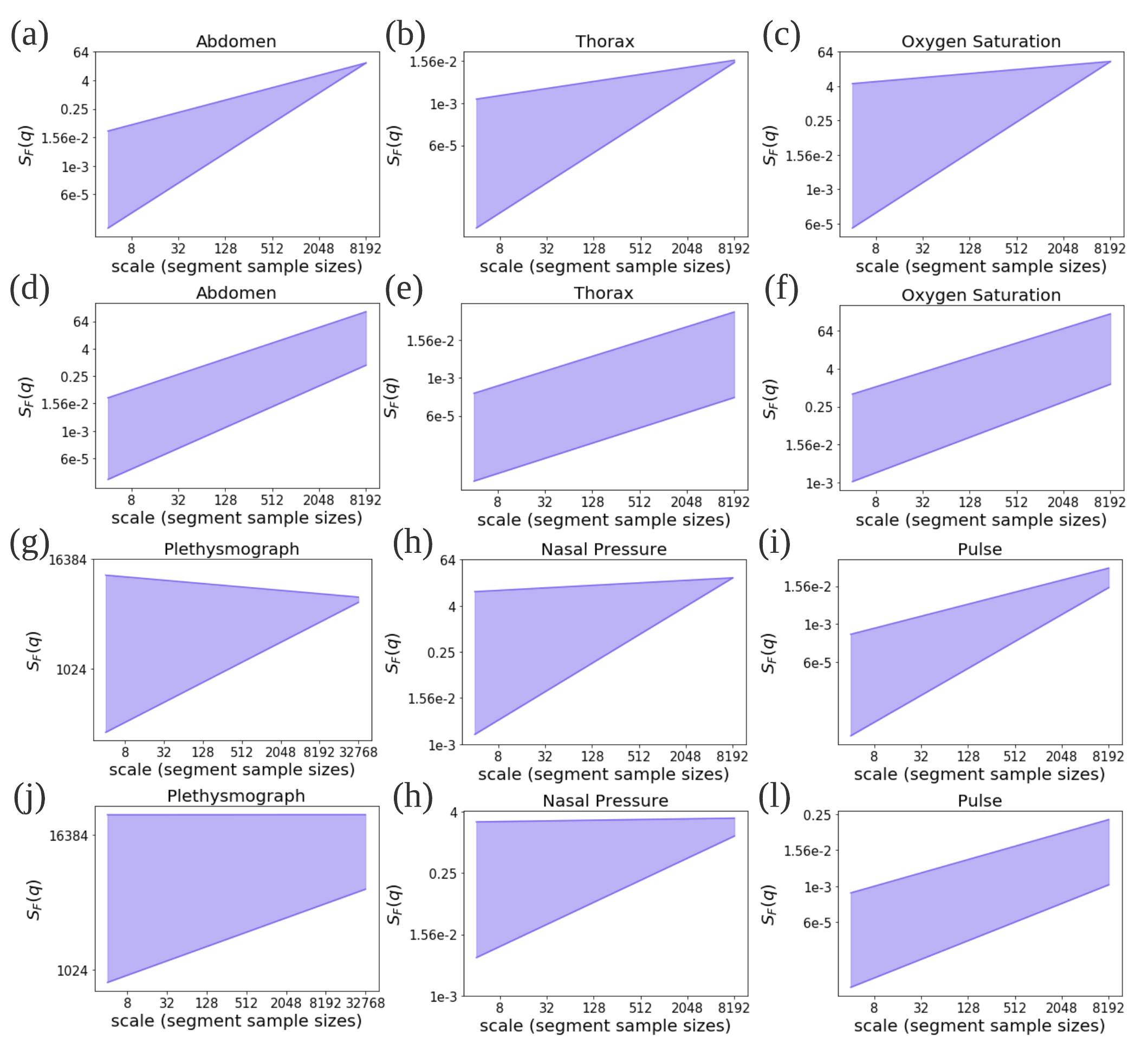} 
\caption{{\textbf{The geometry of fluctuation profiles for the COPD-relevant physiological signals recorded from healthy people and stage 4 patients. } We compute the scaling function from non-derived physiological signals (Abdomen, Thorax, Oxygen Saturation, Plethysmograph, Nasal Pressure, and Pulse) with the exponents $q\in$[-5, 5]. Panels (a-c) and (g-i) correspond to all healthy people in our datasets (stage 0), while (d-f) and (j-l) correspond to all the stage 4 COPD patients in our datasets.}}
\label{fig:hyp_all}
\end{figure}

\begin{figure}[h!]
\centering
\includegraphics[width=\textwidth]{sfigs/mfdfa.png} 
\caption{\response{\textbf{The geometry of fluctuation profiles for the COPD-relevant physiological signals recorded from a normal abdomen and a stage 4 COPD abdomen.} We use the toolbox provided by ~\cite{mukli2015multifractal} to calculate the scaling functions from 6 {raw physiological signals}: Abdomen, Thorax, Oxygen Saturation, Plethysmograph, Nasal Pressure, and Pulse, where the exponents are $q\in$ [-5, 5]. Panels (a-c) and (g-i) with signals recorded from a healthy person, (d-f), and (j-l) with signals recorded from a representative stage 4 COPD patient.}}
\label{fig:mu}
\end{figure}

\begin{figure}
\centering
\includegraphics[width=1\textwidth]{sfigs/datadistribution.png}
\caption{{(a). Data distribution across the medical institutions (b). Percentage of COPD stages for each medical institution}
}
\label{fig:dis}
\end{figure}

\subsection*{\response{Standard deviation analysis and MSE analysis for scaling functions}}
\response{To show the fractal difference between raw signals recorded from stage 0 and 4 participants in detail, we generated Figure~\ref{fig:std} to show the correlations between scale ($s$) vs. standard deviations (STDs) of $S_F(q)$ and $q$ values vs. STDs of $S_F(s)$. In Figure~\ref{fig:std}, Panels (a) and (b) respectively show the $s$ vs. STDs of $S_F(q)$ for stage 0 and 4 participants across 6 physiological signals (abdomen, nasal pressure, oxygen saturation, plethysmograph, pulse, and thorax). When $s=L$, in stage 0, the STD for all six signals is close to zero; in stage 4, the STDs for all signals are larger than 0.5 (except nasal pressure). This observation is also similar to Fig. 1 in Mukli et al. ’s work~\cite{mukli2015multifractal}, where the multifractal signals have smaller STDs of $log(S(q, L))$ than the multifractal noise (monofractal). Accordingly, we can claim that the raw signals across Stage 0 and Stage 4 participants have different fractal features.
To measure the error in terms of scale and fluctuation, we show the mean square error (MSE) of the regressed thin pink lines in Figure 2. We show the results in Figure~\ref{fig:mse}, where (a) and (b) show the MSE for stage 0 and 4 participants, respectively.
}
\begin{figure}
\centering
\includegraphics[width=1\textwidth]{sfigs/revise_1.pdf}
\caption{\response{Scale ($s$) vs. standard deviation (STD) of $S_F(q)$ between stage 0 (a) and stage 4 (b) participants.}
}
\label{fig:std}
\end{figure}
\begin{figure}
\centering
\includegraphics[width=1\textwidth]{sfigs/mse.pdf}
\caption{\response{Mean squared error (MSE) for the regressed lines in Figure 2 for stage 0 and stage 4 participants. }
}
\label{fig:mse}
\end{figure}

\subsection*{{Error analysis and learning process visualization}}

{Although the fractional dynamics deep learning model FDDLM provides relatively high prediction accuracy under $k$-fold cross-validation and hold-out validation, the detection accuracy drops a bit in the hold-out validation (from 98.66\% to 95.88\%). The reason is that the data recorded from each medical institution are unbalanced, as our cohort is a real-life population. The Victor Babes (VB) and Cardio Prevent (CP) are two large clinics with COPD patients who are more willing to get the diagnosis or medical treatment in large units or hospitals than in small clinics—especially in the case of severe and very severe COPD stages. Thus, signals recorded from VB and CP are more comprehensive/diverse than from Medicover 1 (MD1) and Medicover 2 (MD2). In the hold-out section, we balanced the data across different institutions using over-sampling and under-sampling approaches. Indeed, the unbalanced data collection is the leading cause of the prediction accuracy drop in the hold-out section. In Figure~\ref{fig:dis}, we present the data distribution across the medical institutions; in the left-hand side panel, we have the distribution of the absolute number of individuals, while in the right-hand side panel, we present the percentage of COPD stages for each institution. Both panels emphasize the data unbalance problem---MD1 and MD2 have a small number of individuals, and only CP has individuals representing all COPD stages. To handle these imbalanced data across different institutions, we perform over-sampling and under-sampling techniques to guarantee that all institutions have a similar number of testing samples across different COPD stages (e.g., since MD2 does not have stage 4 samples, we move two stage-4 patients from VB to MD2 and over-sample/under-sample the data to ensure the data balance). }

{To provide further insight, we present the learning process in the hidden layers of our fractional dynamics deep learning model for $k$-fold and hold-out validation by employing the t-Distributed Stochastic Neighbor Embedding (t-SNE) visualization algorithm. Besides the input-, dropout-, and output-layers, we have two hidden dense layers in our deep learning model (the first layer has 300 neurons and the second layer has 100 neurons). The t-SNE technique is an approach to reduce data dimensionality in two or three-dimension maps. In this work, we regard the outputs of the first dense layer as a 300-dimension coordinate and the second dense layer as a 100-dimension coordinate. Then, we employ the t-SNE technology to reduce these coordinates to two-dimension coordinates and visualize the learning processes in these two hidden dense layers. Figure~\ref{fig:tsne1} presents the learning processes visualization of the first hidden layer. In Figure~\ref{fig:tsne1}, panels (a-d) present the t-SNE results for hold-out validation, and panels (e-i) show the t-SNE results for $k$-fold cross-validation. The color of scatters presents the label of each individual case, and each point's shape indicates the prediction results of our fractional dynamics deep learning model (where the round markers indicate the correct prediction results and the 'x' markers indicate errors). Figure~\ref{fig:tsne1} (a-d) hold out the data recorded from different institutions as test sets (VB (a), MD1 (b), MD2 (c), and CP (d)), and (e-i) hold out the 0-20\% (e), 20\%-40\% (f), 40\%-60\% (g), 60\%-80\% (h), 80\%-100\% (i) data as test sets, respectively. In Figure~\ref{fig:tsne1}, we can find that scatters under the same category tend to build clusters, and some clusters overlap. The overlapping patterns are difficult to learn for the deep learning model (errors mostly occur in these overlapping ranges), and the isolated clusters with clear boundaries are the easiest part to recognize for the deep learning model. By comparing the t-SNE figures across hold-out validation (Figure~\ref{fig:tsne1} (a-d)) and $k$-fold cross-validation (Figure~\ref{fig:tsne1} (e-i)), we find that the scatters in hold-out validation have more extensive overlapping ranges than the scatters in $k$-fold cross-validation. This observation illustrates that the datasets used in hold-out validation are harder to distinguish for the deep learning model than the dataset used in $k$-fold cross-validation.}

{Figure~\ref{fig:tsne2} presents the learning processes visualization of the last hidden layer in our fractional dynamics deep learning model. Figure~\ref{fig:tsne2} (a-d) hold out the data gathered from different institutions as test sets (VB (a), MD1 (b), MD2 (c), and CP (d)), and (e-i) hold out the 0-20\% (e), 20\%-40\% (f), 40\%-60\% (g), 60\%-80\% (h), 80\%-100\% (i) data as test sets, respectively. As one can observe, Figure~\ref{fig:tsne2} has the same occurrence as in Figure~\ref{fig:tsne1}; namely, that the datasets used in hold-out validation are more difficult to learn than the dataset used in $k$-fold cross-validation. Moreover, by comparing Figure~\ref{fig:tsne1} and Figure~\ref{fig:tsne2}, we observe that overlapping patterns in each Figure~\ref{fig:tsne2} panel are narrower than the corresponding panel in Figure~\ref{fig:tsne1}. This provides numerical evidence that our fractional dynamics deep learning model FDDLM provides a feasible learning process and gives desired prediction results. }
\begin{figure}
\centering
\includegraphics[width=1\textwidth]{sfigs/tsne_layer1.pdf}
\caption{{\textbf{Representation of the learning process about hold-out and $k$-fold cross-validation in the first hidden layer.} We present the t-SNE visualization of the first hidden layer in the fractional dynamics deep learning model of the COPD dataset under hold-out (a-d) and $k$-fold (e-i) cross-validation. In each scatterplot, the colors of the points indicate the classes of the corresponding COPD stages, and the shape of the points represents the fractional dynamics deep learning model’s prediction results (i.e., the round shape indicates a correct prediction, and the ‘x’  mark indicates an error). }
}
\label{fig:tsne1}
\end{figure}

\begin{figure}
\centering
\includegraphics[width=1\textwidth]{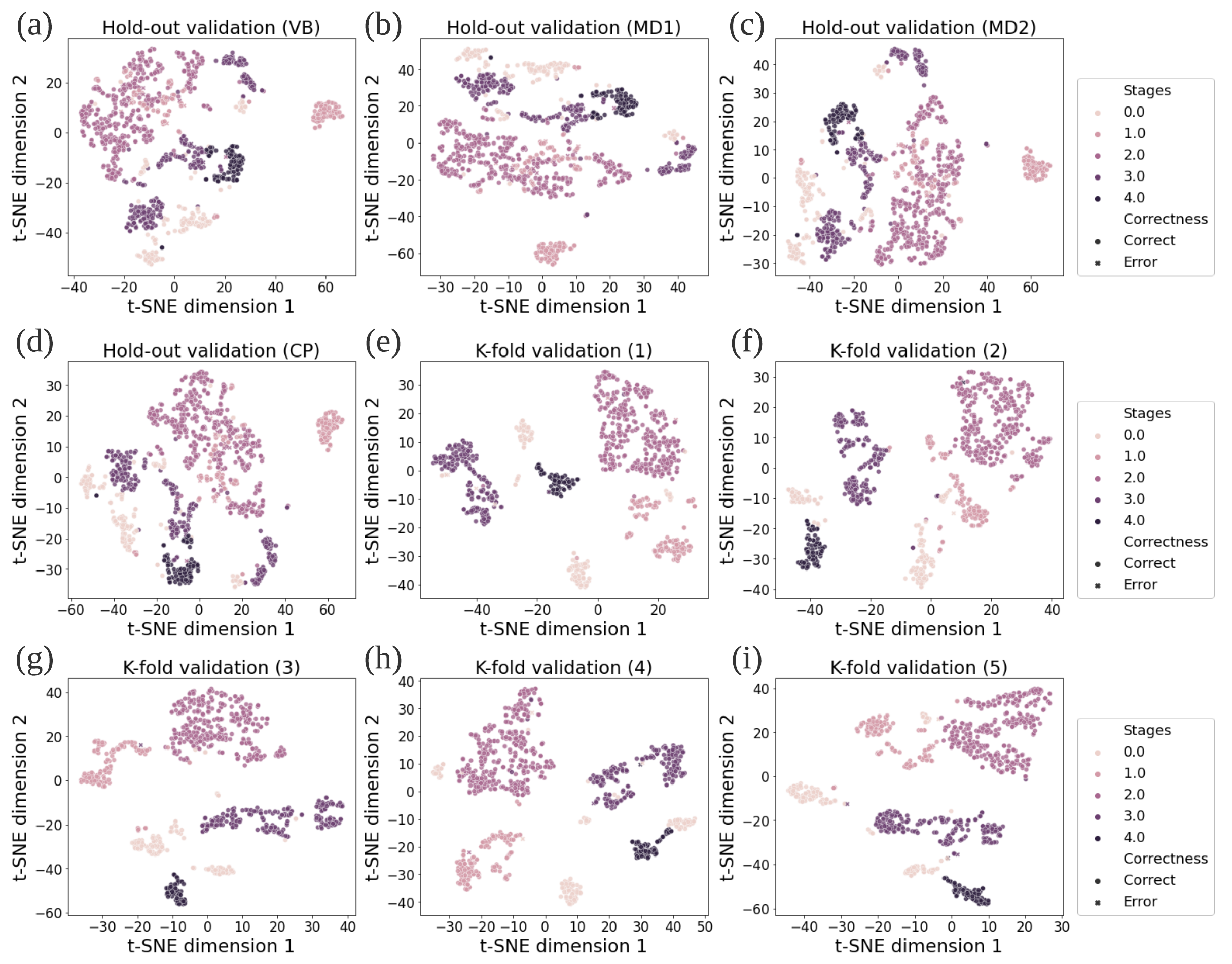}
\caption{{\textbf{Representation of the learning process about hold-out and $k$-fold cross-validation in the last hidden layer.} We present the t-SNE visualization of the last hidden layer in the fractional dynamics deep learning model of the COPD dataset under hold-out (a-d) and $k$-fold (e-i) cross-validation. In each scatterplot, the colors of the points indicate the classes of the corresponding COPD stages, and the shape of the points represents the fractional dynamics deep learning model’s prediction results (i.e., the round shape indicates a correct prediction, and the ‘x’  mark indicates an error).}}
\label{fig:tsne2}
\end{figure}

\subsection*{{Linear classifier of coupling matrices}}
{Besides the deep learning architecture in our fractional dynamics deep learning model FDDLM, we also utilize linear classifier models to mine the complexity of coupling matrices $A$. Logistic regression is traditionally a linear classifier; it is one of the most used two-class (and multi-class) classification machine learning algorithms\cite{hosmer2013applied}. When we use logistic regression instead of deep learning on the $A$ matrices, the accuracy of COPD prediction is $94.61\%\pm0.98\%$ with $k$-fold and $86.68\%\pm0.82\%$ with hold-out validation. The COPD prediction accuracy for logistic regression is lower with the $k$-fold validation and significantly lower with the hold-out validation than the fractional dynamics deep learning model (i.e., $98.66\%\pm0.45\%$ and $95.88\%\pm1.76\%$, respectively). We show the confusion matrices for logistic regression in Figure~\ref{fig:log}; panels (a-e) are the prediction results on test sets for the $k$-fold validation, and (f-i) panels show the prediction results for the hold-out validation. Of note, for the linear classifier model, the accuracy for the $k$-fold validation is significantly higher than the accuracy for hold-out validation. The explanation is that our data is unbalanced in different medical institutions (for detailed information, see the \emph{Supplementary material}'s section \emph{Error analysis and learning process visualization}).}

{Besides the logistic regression, we also investigate another classifier model as a reference, namely, SVM. Support-vector machines (SVM) are a particular case of linear classifiers based on the margin maximization principle\cite{joachims2006training}. When using SVMs instead of deep learning on the $A$ matrices, the accuracy of COPD prediction becomes $94.74\%\pm0.24\%$ with $k$-fold and $92.62\%\pm0.39\%$ with hold-out validation. We show the confusion matrices for SVM under $k$-fold and hold-out validations in Figure~\ref{fig:svm}. In Figure~\ref{fig:svm}, panels (a-e) are the prediction results on test sets for the $k$-fold validation, and (f-i) panels show the prediction results for the hold-out validation. Like the logistic regression model, the $k$-fold validation accuracy is significantly higher than the accuracy for hold-out validation in the SVM model. The reason is that our data is unbalanced across the institutions. We conclude that our fractional dynamics deep learning model outperforms these two linear classifier models (i.e., logistic regression and SVM) in terms of prediction accuracy under $k$-fold and hold-out validation. Indeed, learning from many matrices $A$ is a complicated process that requires deep learning.}

\begin{figure}[h!]
\centering
\includegraphics[width=1\textwidth]{sfigs/linear_classify.pdf}
\caption{{\textbf{The confusion matrices resulted from the logistic regression model.} The $k$-fold cross validation ($k$=5): in each iteration, we hold out  0-20\% (a), 20\%-40\% (b), 40\%-60\% (c), 60\%-80\% (d), and 80\%-100\% data (e) as test sets. For the hold-out validation, we built the test sets by holding out data recorded from one institution at a time: VB (f), MD1 (g), MD2 (h), and CP (i).  
}}
\label{fig:log}
\end{figure}

\begin{figure}[h!]
\centering
\includegraphics[width=1\textwidth]{sfigs/svm.pdf}
\caption{{\textbf{The confusion matrices resulted for the SVM.} For the $k$-fold cross validation ($k$=5): in each iteration, we hold out  0-20\% (a), 20\%-40\% (b), 40\%-60\% (c), 60\%-80\% (d), and 80\%-100\% data (e) as test sets. For the hold-out validation, we built the test sets by holding out data recorded from one institution at a time: VB (f), MD1 (g), MD2 (h), and CP (i).  
}}
\label{fig:svm}
\end{figure}

\subsection*{{Standard questionnaires, exacerbation history, and comorbidities of COPD patients}}

{In this section, we provide the data description about the standard questionnaires (CAT and MRC), exacerbation history, and comorbidities about all the COPD patients in our datasets. Questionnaires are recommended for the management of COPD. Since 2011, the GOLD guidelines have included the following questionnaires in the assessment of COPD patients: the modified Medical Research Council (mMRC) dyspnea scale\cite{bestall1999usefulness}, the COPD assessment test (CAT)\cite{jones2009development}, and the clinical COPD questionnaire (CCQ); they are carefully-designed high-quality questionnaires, but information on the feasibility for routine use is scarce. Nonetheless, questionnaires are both quick to complete and have good acceptability by the patient. In addition, the agreement between electronic and paper versions of the questionnaires was high. Figure~\ref{fig:cat} presents the CAT and MRC scores for COPD patients across different stages in our paper's WestRo cohort. However, we made sure that our diagnoses---that lead to classifying patients in COPD stages 0 to 4---are reliable by reviewing each case after several months (consisting of a complete medical check-up, including spirometry). }
\begin{figure}[h]
\centering
\includegraphics[width=1\textwidth]{sfigs/catscore.png}
\caption{{CAT (a) and MRC (b) scores for COPD patients across different COPD stages}
}
\label{fig:cat}
\end{figure}

\begin{figure}
\centering
\includegraphics[width=1\textwidth]{sfigs/exacerbations.png}
\caption{{(a) Exacerbations across patients under different COPD stages (b) Percentage of COPD stages for each comorbidity (Cardiovascular comorbidities (CC), Cancers (CA), Metabolic comorbidities (MC), Psychiatric comorbidities (PC), Renal disease (RD)). }}
\label{fig:ex}
\end{figure}

\begin{figure}[h]
\centering
\includegraphics[width=1\textwidth]{sfigs/conver.pdf}
\caption{{Wasserstein distance between the coupling matrices $A$ over time. }}
\label{fig:con}
\end{figure}

{We also collected data about the history of exacerbation for all the patients. When they were evaluated, their condition was stable. The data about the exacerbation history in the previous year for our patients is 16 patients without exacerbations, 25 patients with one exacerbation, 5 patients with two exacerbations, and 1 with three exacerbations. We considered it essential to make all the measurements and evaluations in a stable COPD phase of the disease. Therefore, we performed all functional tests when patients were not experiencing an acute exacerbation. The reason is that there is an increase in hyperinflation and gas trapping during an exacerbation, with a reduced expiratory flow and increased dyspnea~\cite{halpin2021global}.}

{Exacerbations are important in the management of COPD patients. There is an \emph{frequent exacerbator} phenotype with an increased risk of hospitalization and death\cite{wedzicha2013mechanisms}. Exacerbations of COPD have a considerable impact on patients’ health status and exercise capacity and have a cumulative effect on lung function~\cite{wedzicha2013mechanisms}. However, longitudinal changes in FEV1 are not significantly associated with the exacerbation risk. Exacerbations can be found in any COPD stage~\cite{jo2019longitudinal}. In addition, a single COPD exacerbation may also result in a significant increase in lung function decline rate~\cite{halpin2017effect}. To investigate the impact of exacerbations, we pick 4 patients from each COPD stage (i.e., stage 2, 3, and 4) and hold out untrained signal samples as test sets (where 2 of them have 2 or more exacerbations and the other 2 have less than 2 exacerbations). We apply our model to make predictions about these test sets to reveal whether the exacerbation history will influence our prediction accuracy (the reason for choosing 4 patients in each stage from our datasets is that we want to maintain the test sets balanced during the prediction process). In stage 2, the prediction accuracy of samples gathered from patients with less than 2 exacerbations (category 1) is $99.01\%$ and the prediction accuracy of samples generated from patients with more than 2 exacerbations (category 2) is $98.07\%$. In stage 3, the prediction accuracy is $99.28\%$ for category 1 and $98.23\%$ for category 2. In stage 4, the prediction accuracy of category 1 is $97.88\%$, and of category 2 is $98.22\%$. Hence, we conclude that the distribution of exacerbations history and comorbidities across patients and stages correlated with the high prediction rate of our method suggests that our fractal dynamics deep learning model FDDLM is not influenced by comorbidities or exacerbation history. (We have patients with exacerbation history in all COPD stages, except—of course—stage 0 COPD; we also have many COPD stages represented in each comorbidity, please see Figure~\ref{fig:ex}.)}

Comorbidities in COPD are expected at any stage of the disease ~\cite{agusti2010characterisation}. The most common comorbidities accompanying COPD include cardiovascular diseases, metabolic disorders, osteoporosis, musculo-skeletal dysfunction, anxiety/depression, cognitive impairment, gastrointestinal diseases, and respiratory conditions such as asthma, bronchiectasis, pulmonary fibrosis, and lung cancer~\cite{divo2012comorbidities}. Comorbidities are known to pose a challenge in the assessment and effective management of COPD. However, the mechanistic links between COPD and its comorbidities are still not fully understood. The variability of the clinical presentation in COPD interacts with comorbidities to form a complex clinical scenario for clinicians to deal with~\cite{lopez2020integrating}. As a result, attention needs to be paid to assessing and managing comorbidities in COPD in both clinical and research settings. In addition, for the effective management of comorbidities in COPD, there is a need for reliable measurement tools that can assist in improving clinical outcomes. In this work, we record the following comorbidities: Cardiovascular comorbidities (CC); Cancers (CA); Metabolic comorbidities (MC); Psychiatric comorbidities (PC); and Renal disease (RD). The results are shown in Figure~\ref{fig:ex} (b). 

\subsection*{{Convergence of coupling matrix}}
{To evaluate our fractional dynamics deep learning model's efficiency, we investigate the sufficient length of the signals to make stable conclusive results. Figure~\ref{fig:con} displays the convergence of matrix $A$ across different stages of patients. The Wasserstein distance is a metric for the differences between two distributions (the small distance value implies that two distributions are similar.) In Figure~\ref{fig:con}, we calculate the coupling matrices for different patients with different COPD stages by the length of the physiological signals (time-series) and calculate the Wasserstein distances between two neighboring time stamps (e.g., we calculate the Wasserstein distance between two coupling matrices generated from time-sequence $0$ to $t$ seconds and $0$ to $t+1$ seconds). From Figure~\ref{fig:con}, we observed that—for each COPD stage—the Wasserstein distance of coupling matrices converges to a small value (less than 0.02) after 600 seconds. This observation suggests that after 10 minutes, the coupling matrices' elements remain constant. Thus, a 10 minutes record interval is sufficient for our model to make the predictions.}

\subsection*{{Early Viral infection detection}}

{To emphasize the generality and correctness of our fractional dynamics method, we also analyze the Biochronicity viral prediction dataset with our model. The Biochronicity pilot study---relevant to the current work---is summarized as follows. The human rhinovirus (HRV)39 was injected into 18 human subjects monitored from 4 days before to 4 days after injection. Out of the 18 patients, 11 were shedding, showed symptoms after 3-4 days, and were marked as infected; the remaining 7 were considered healthy. The E4 Empatica physiological sensor records the temperature, blood pressure-volume (heart rate), accelerometer (3-axis), skin temperature, and electrodermal activity. The goal was to detect the viral infection in less than 24hrs from the inoculation point. The dataset imposes several practical challenges: first, paucity of data as there are only 18 samples; second, the detection has to be made as early as 24 hrs after the viral inoculation---when the symptoms are not prominent enough to be detected by a medical expert.}

{Next, we explain the fractional dynamics-based method/pipeline in viral prediction. We take the 3 physiological time-series features, electrodermal activity (EDA), body temperature (TEMP), and inter-beat interval (IBI). For a given inoculation point, the 3-dimensional time series are broken into pre- and post-viral infection data. For each pre- and post-viral data, a sliding window mechanism with a window length of 3000 samples and a sliding length of 100 samples (the choice is made by cross-validation over the window length and sliding length grid) is fitted using a fractional dynamical model with spatial coupling, from which we obtain the fractional coefficients $\alpha$. Each window slide results in three fractional coefficients (one for each physiological feature), and then we estimate the probability density of $\alpha$ for pre- and post-viral periods. Finally, we use the Kullback-Leibler (KL) divergence between the pre- and post-viral fractional distributions as the feature for differentiating the infected and healthy subjects. The intuition is that the fractional coefficient captures the scaling behavior of the time series; by computing the difference between the distributions, we assume that healthy and infected subjects have different scaling behavior in their physiological activities.}

\begin{figure}
\vspace{-10pt}
\includegraphics[width=\linewidth]{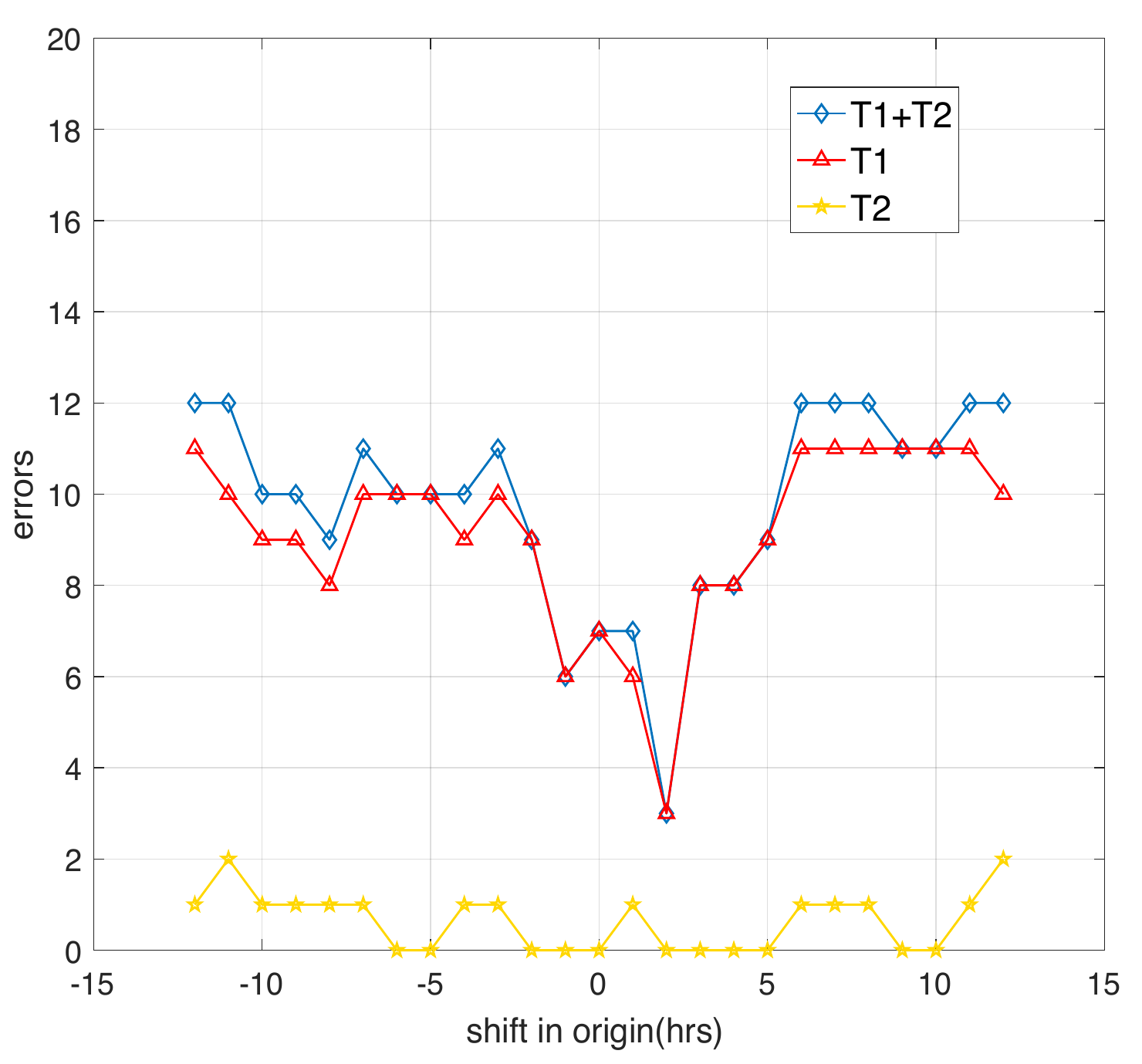}
\caption{{Early viral prediction using fractional dynamics. The Type-I and Type-II errors are shown by shifting the model's viral injection point from the actual reference (shift=0).}}
\label{fig:viral}
\end{figure}

{One of the crucial assumptions in early viral detection is the knowledge of the viral inoculation point. From a practical standpoint, it is not possible to obtain such information. Therefore, we evaluate the efficacy of our model by moving the assumed viral inoculation point from the actual infection time in both directions. Figure\,\ref{fig:viral} shows the plot of classification performance in terms of Type-I and Type-II errors when the inoculation point is shifted in positive and negative increments. Due to the scarcity of data, we use a leave-one-out-based resampling for the classification task. The ‘V’-like structure of the Type-I error is informative because it shows the best performance (as few as 3 errors) close to the actual inoculation point (0 hrs) while the loss in performance on moving in either direction from 0 hours. The error increase (in either direction) is caused by shifting the inoculation point; the past and future of the patient become more or less the same. Hence, the classifier would falsely label the patient as healthy or, in other words, make a Type-I error. Another important observation is that the best performance---a total of 3 errors out of 18 subjects (3 Type-I and 0 Type-II)---is achieved not at the actual infection time (shift = 0 hours) but when the infection time is assumed delayed by +1 hour. A possible explanation is that the physiological effects take time to trigger; therefore, a delayed version of the separating boundary produces better differentiation between the pre- and post-viral data. Finally, another important outcome of the fractional model is that the Type-II error is roughly constant and is consistently low, which is required behavior.}

\begin{figure}[h!]
\centering
\includegraphics[width=1\textwidth]{sfigs/fig_new.pdf}
\caption{{\textbf{Training and testing result comparisons of different deep learning models (WestRo Porti dataset) for the $k$-fold cross-validation ($k=5$).} The training/testing accuracy (a), loss (b), and AUC (c) for our fractional dynamics deep learning approach, where the training processes use signal signatures extracted with the fractional dynamic mathematical model. The training/testing accuracy (d), loss (e), and AUC (f) for the Vanilla DNN model, where the training processes use the physiological signals recorded with the Porti portable sleep monitors. The training/testing accuracy (g), loss (h), AUC (f) for the LSTM model, where the training processes use the physiological signals. The training/testing accuracy (j), loss (k), and AUC (l) for the CNN model, where the models are trained with physiological signals.}}
\label{fig:k_fold_new}
\end{figure}

\subsection*{WestRo Porti COPD dataset analysis}

To test the generalization capabilities of our framework, we use the WestRo Porti COPD dataset consisting of 13824 physiological signals samples, recorded from 534 patients (232 COPD patients and 302 non-COPD patients) in the Victor Babes hospital. The dataset recorded 6 physiological signals from each patients (for detailed information about WestRo Porti COPD dataset, see the \emph{Method} section \emph{Data collection}). This section evaluates the accuracy, loss, and area under the curve (AUC). The results generated from different models (fractional-dynamics deep learning model, Vanilla DNN, LSTM, and CNN) are validated using the $k$-fold cross-validation approach (where $k=5$). Figures~\ref{fig:k_fold_new} (a), (b), and (c) present our fractional-dynamics deep learning model's performance in terms of accuracy, loss, and AUC among training and testing datasets, respectively. Figure~\ref{fig:k_fold_new} (a) shows that both training and testing accuracy curves present an increasing trend---suggesting that our model improves the prediction accuracy over time without overfitting---to a predicting accuracy of $96.18\%\pm0.48\%$. Figure~\ref{fig:k_fold_new} (b) illustrates that with increasing of the epoch numbers, the training and testing AUC curves also converge to optimal steady states. Figure~\ref{fig:k_fold_new} (c) shows that the loss curves for both training and testing processes exhibit a decreasing tendency, which indicates that, in our neural network, the value of the loss function converges over epochs.

Figure~\ref{fig:k_fold_new} (d), (e), and (f), respectively, show the training and testing accuracy, loss, and AUC curves obtained from the Vanilla DNN model trained with physiological signals (raw data). The testing results for the Vanilla DNN model show that the testing accuracy and AUC curves present a decreasing trend over epochs, which is evidence for overfitting. Consequently, we employ the early-stop mechanism to maintain the performance of the Vanilla DNN model (save the best-performance model). The best-performance model under Vanilla DNN exhibits a much lower prediction accuracy ($31.23\%\pm4.58\%$) than the fractional dynamics deep learning model, which illustrates that our model outperforms the Vanilla DNN model.

Figure~\ref{fig:k_fold_new} (g), (h), and (i), respectively, present the training and testing accuracy, loss, and AUC results curves for the LSTM model trained with physiological signals (raw data). Observing the training and testing result curves in Figure~\ref{fig:k_fold_new} (g-i), we notice that the LSTM model also overfits; to deal with this situation, we also utilize the early-stopping mechanism to choose the best-performance LSTM model. The prediction accuracy of the best-performance LSTM model yielded $81.17\%\pm2.52\%$, whereas our fractional dynamics deep learning model presents a significantly higher prediction accuracy of $96.18\%\pm0.48\%$.

{We also investigated whether the convolutional neural network (CNN) model can outperform our fractional-dynamics deep learning model by characterizing the dynamics of the physiological signals with higher accuracy. The results are presented in Figure~\ref{fig:k_fold_new} (j-l). Figure~\ref{fig:k_fold_new} (j), (h), and (i), respectively, show the training and testing accuracy, loss, and AUC results for the CNN model trained with physiological signals (raw data). The results show that the CNN model correctly classified $23.77\%\pm0.02\%$ COPD samples under $k$-fold cross-validation ($k=5$). Thus, our fractional dynamics deep learning model predicts patients' COPD stages with a much higher accuracy than the Vanilla DNN, LSTM, and CNN models---trained with physiological signals (raw data)---without overfitting (for detailed information about network architecture, see section \emph{Methods}, subsection \emph{Neural network architecture for the WestRo Porti COPD dataset}). Of note, the testing accuracy of $96.18\%\pm0.48\%$ for the WestRo Porti COPD dataset is slightly lower than the testing accuracy for the WestRo COPD dataset. The reason is that the coupling matrix for the WestRo COPD dataset has more features than the coupling matrix for the WestRo Porti COPD dataset (i.e., 144 features vs. 36 features), and fewer features degrade the predicting accuracy.}

\begin{figure}[h!]
\centering
\includegraphics[width=1\textwidth]{sfigs/acc_new.pdf}
\caption{{\textbf{Training and testing result comparisons of different deep learning models (WestRo Porti dataset) for the $k$-fold cross-validation ($k=5$).} The training/testing accuracy (a), loss (b), and AUC (c) for our fractional dynamics deep learning approach, where the training processes use signal signatures extracted with the fractional dynamic mathematical model. The training/testing accuracy (d), loss (e), and AUC (f) for the Vanilla DNN model, where the training processes use the physiological signals recorded with the Porti portable sleep monitors. The training/testing accuracy (g), loss (h), AUC (f) for the LSTM model, where the training processes use the physiological signals. The training/testing accuracy (j), loss (k), and AUC (l) for the CNN model, where the models are trained with physiological signals.}}
\label{fig:patient}
\end{figure}

\subsection*{Pre-patient based $K$-fold analysis}
In this section, we performed the FDDLM's k-fold cross-validation results such that training does not use data from individuals considered in testing. (Nonetheless, our hold-out validation makes this type of evaluation implicit because each patient belongs to just one institution). Fig. \ref{fig:patient} presents our model’s performance as accuracy, loss, and AUC curves between training and testing datasets (where $k = 5$). Based on Fig. \ref{fig:patient}, we find that our model yielded an accuracy of  $98.70\% \pm 0.407\%$ (very close to the accuracy we reported in the initially submitted manuscript).

\bibliography{sample}